\documentclass[3p,final,11pt,authoryear]{elsarticle}
\usepackage[colorlinks]{hyperref}
\usepackage[utf8]{inputenc}
\usepackage{graphicx}
\usepackage{mathtools}
\usepackage{amsmath}
\usepackage{lscape}
\usepackage[nameinlink,noabbrev]{cleveref}
\usepackage{enumitem}
\usepackage{natbib}
\usepackage{subcaption}
\usepackage[tableposition = top]{caption}
\usepackage{booktabs}
\usepackage{algorithm}
\usepackage{algorithmic}
\usepackage{hyperref}
\usepackage{float}
\usepackage{tabularx}
\usepackage{adjustbox}
\usepackage{rotating}
\usepackage{lscape}
\usepackage{amsfonts}

\setcitestyle{authoryear,open={(},close={)}}
\usepackage{setspace}

\begin{document}

\begin{frontmatter}

\title{Towards Locally Deployable Fine-Tuned Causal Large Language Models for Mode Choice Behaviour} 

    \author[1]{Tareq Alsaleh\corref{cor1}}\ead{talsaleh@torontomu.ca}
    \author[1]{Bilal Farooq}\ead{bilal.farooq@torontomu.ca}

    \address[1]{Laboratory of Innovations in Transportation (LiTrans), Toronto Metropolitan University, Canada}
    \cortext[cor1]{Corresponding Author.}

\begin{abstract}
This study investigates the adoption of open-access, locally deployable causal large language models (LLMs) for travel mode choice prediction and introduces LiTransMC, the first fine-tuned causal LLM developed for this task. We systematically benchmark eleven open-access LLMs (1-12B parameters) across three stated and revealed preference datasets, testing 396 configurations and generating over 79,000 mode choice decisions. Beyond predictive accuracy, we evaluate models generated reasoning using BERTopic for topic modelling and a novel Explanation Strength Index, providing the first structured analysis of how LLMs articulate decision factors in alignment with behavioural theory.
LiTransMC, fine-tuned using parameter efficient and loss masking strategy, achieved a weighted F1 score of 0.6845 and a Jensen-Shannon Divergence of 0.000245, surpassing both untuned local models and larger proprietary systems, including GPT-4o with advanced persona inference and embedding-based loading, while also outperforming classical mode choice methods such as discrete choice models and machine learning classifiers for the same dataset. This dual improvement, i.e., high instant-level accuracy and near-perfect distributional calibration, demonstrates the feasibility of creating specialist, locally deployable LLMs that integrate prediction and interpretability.
Through combining structured behavioural prediction with natural language reasoning, this work unlocks the potential for conversational, multi-task transport models capable of supporting agent-based simulations, policy testing, and behavioural insight generation. These findings establish a pathway for transforming general purpose LLMs into specialized and explainable tools for transportation research and policy formulation, while maintaining privacy, reducing cost, and broadening access through local deployment.

\end{abstract}

\end{frontmatter}

\section{Introduction}
\label{sec:Introduction}
Traditional transportation modelling has evolved through several methodological advances, each building upon the previous generation, yet leaving some important gaps. The random utility maximization framework introduced by McFadden \citep{mcfadden1974measurement} and formalized in Ben-Akiva and Lerman's seminal textbook \citep{ben1985discrete} established discrete choice models as the foundation of travel demand analysis. These models provided behavioural interpretability, but relied on restrictive assumptions such as independence of irrelevant alternatives (IIA) and linear-in-parameters utility functions, which constrained their ability to capture complex substitution patterns. To overcome some of these restrictions, simulation-based estimation methods were introduced \citep{train2009discrete}, and hybrid choice models incorporated attitudinal and perceptual constructs \citep{ben2002integration}. While these extensions enhanced behavioural realism, they still depended on structured survey variables and struggled to integrate unstructured or contextual information that influences traveller decisions.

Parallel explorations in machine learning (ML) offered another direction. Neural networks and related classifiers \citep{shmueli1996neural, hensher2000comparison} demonstrated higher predictive accuracy by capturing non-linearities and interactions in addition to their ability to incorporate other ubiquitous data, like gps trajectories and WiFi signals for choice modelling. However, ML models typically sacrificed interpretability and departed from established behavioural theory. In practice, this meant that while econometric models could explain why a decision was made but sometimes lacked predictive power, machine learning models could predict choices without offering transparent behavioural insights. Moreover, the practical execution of travel surveys is increasingly hindered by declining response rates and recruitment difficulties, which undermine the representativeness of samples and introduce nonresponse bias in ways that are difficult to correct \citep{wittwer2024new, wang2023response, svaboe2024comparative}.

This growing debate between behavioural realism and predictive power combined with lower participation rates sets the stage for advances from Large Language Models (LLMs), which provide the capability to process natural language, incorporate qualitative context, and articulate reasoning in ways that earlier models could not. Additionally, their capacity for few-shot generalization enables the inference of behavioural regularities from limited examples, which supports the creation of data-efficient digital representations of respondents that capture contextual and cognitive diversity beyond what conventional surveys can offer. In natural language processing, early recurrent and LSTM-based models \citep{hochreiter1997long, mikolov2010recurrent} improved over statistical n-grams, but the real breakthrough came with the transformer architecture \citep{vaswani2017attention}. This enabled large-scale pretraining, exemplified by BERT's bidirectional encoder \citep{devlin2019bert} and the GPT series of autoregressive decoders \citep{brown2020language}. Subsequent work on scaling laws \citep{kaplan2020scaling, hoffmann2022training}, retrieval-augmented generation \citep{borgeaud2022improving}, and alignment methods such as the Reinforcement Learning with Human Feedback (RLHF) \citep{ouyang2022training} shaped the trajectory toward today's frontier models, including closed systems like the proprietary GPT series \citep{achiam2023gpt} and open-source alternatives such as LLaMA \citep{touvron2023llama}.

The recent surge of interest in generative artificial intelligence has seen LLMs applied across a widening spectrum of domains. These models are trained on vast corpora of text and demonstrate emergent abilities in contextual understanding, multi‑modal data fusion and reasoning, offering a new way for processing complex, human centred information \citep{zhang2025survey}. In transportation research, early commentaries by \cite{mahmud2025} highlighted how LLMs could support intelligent transportation systems (ITS) by predicting traffic flow, detecting vehicles and assisting autonomous driving. Subsequent surveys and frameworks by \cite{nie2025} categorize the roles of LLMs as information processors, knowledge encoders, component generators and decision facilitators, while \cite{zhang2025survey, mdpi2024review} illustrate their potential for forecasting mobility time series, where they integrate unstructured textual inputs and enable human–machine interaction. At the same time, researchers caution that proprietary LLMs carry high computational costs, and privacy risks; open‑source models and published fine‑tuned derivatives are recommended to ensure transparency, reproducibility and reduced cost \citep{mdpi2024review}.

Driven by these opportunities, several studies have begun to explore LLMs for mobility analysis. \cite{wang2023would} introduced LLM‑Mob, a framework that reorganizes mobility data into historical and context stays and designs context inclusive prompts so that a general purpose GPT‑3.5 model can predict a person’s next location.  Their results showed that careful prompt engineering enables LLMs to capture both long‑ and short‑term dependencies and produce accurate, interpretable predictions.  \cite{guo2024towards} proposed xTP‑LLM for traffic flow prediction, converting multi‑modal traffic data into natural language descriptions and fine‑tuning a LLaMA‑based model with instruction tuning. Their approach achieved competitive accuracy while offering intuitive textual explanations, marking one of the first applications of LLMs to spatio‑temporal traffic forecasting. Other works leverage LLMs as world simulators and assistants \cite{li2024mobagent} presented MobAgent, an agent‑based framework that uses GPT‑4 and DeepSeek APIs to extract mobility patterns and recursively reason about individual motivations in order to generate realistic, personalised travel diaries. These generative applications highlight the promise of LLMs for producing human‑centred narratives. 

More closely related to travel behaviour and mode choice prediction modelling, \cite{mo2023large} proposed using LLMs to predict travel mode choice in a zero‑shot setting. They design prompts that include task description, individual attributes and domain knowledge and ask ChatGPT 3.5 turbo to predict a commuter’s mode choice without any training data. Their results reveal that LLM based predictions can attain accuracy similar to classical discrete choice models and machine‑learning classifiers, and that LLMs can articulate the reasoning behind decisions. However, the authors also observe occasional logical violations and hallucinations in the generated explanations. This highlights how granting LLMs access to open-access datasets and domain-specific research corpora could substantially enhance their capacity to acquire specialised transport knowledge, moving beyond the limitations of general-purpose pre-training. In \cite{liu2025aligning} work, we do not see a replication of the accuracy at the zero-shot level, but the authors build on this work and propose a framework for alignment using a persona‑based embedding learning approach through ChatGPT 4o for few-shot prompting. They infer personas from socio‑demographic groups through representation learning and condition prompts on these personas. This persona loading significantly improves the match between LLM predictions and human travel choices while remaining computationally tractable.  Despite these advances, both studies rely on general‑purpose proprietary models accessed via APIs. No fine‑tuned models specialized for travel choice exist, and there is little discussion of privacy or deployment cost.  Moreover, these studies do not provide a systematic analysis of how the textual reasoning aligns with behavioural theory beyond anecdotal observations.

Outside of travel behaviour modelling, some efforts explore the potential of fine-tuned LLMs on transportation safety applications. For example, \cite{mdpi2024review} fine‑tuned the LlaMA‑7B model on the TrafficSafety‑2K dataset to produce TrafficSafetyGPT, which outperformed the base LlaMA on transportation safety tasks. The work that comes closest to our own is by \cite{bhandari2024urban}, who also fine-tuned an open-source model for transport behaviour. In their study, they fine-tuned a Llama-2-7B model on household travel survey data, specifically training it to generate synthetic multi-day travel diaries. The model learned the sequential patterns of human mobility, producing entire schedules of activities and trips. The crucial distinction lies in the nature of the task and the research objective. This work addresses a complex, generative challenge of sequence creation, aiming to replicate the statistical distribution of entire travel patterns for applications like large-scale agent-based simulations. Our research, conversely, tackles a targeted predictive challenge,  determining the travel mode for a single trip given a specific context. By isolating the mode choice decision, our approach allows for direct performance validation against established discrete choice models and creates a specialized tool for scenario-based policy analysis. 

The lack of fine‑tuned LLMs for travel mode choice prediction indicates a significant research gap. A comprehensive review also urges researchers to prefer open‑source models and to publish fine‑tuned derivatives to facilitate reproducibility \citep{mdpi2024review}.
Despite a growing interest, research on LLMs in transportation remains limited. Most studies have focused on specific use cases or a single LLM model or framework, restricting the generalizability of findings across diverse contexts. High computational costs often hinder research efforts, while sensitive or proprietary datasets remain inaccessible to many researchers. Data privacy concerns are particularly significant when dealing with personal mobility data, which highlights the need for a comprehensive testing framework. Such a framework should evaluate open-access, computationally feasible LLMs, detailing their potential and limitations across various applications.

Building on these insights, our study advances travel choice modelling with three key contributions.  

\begin{itemize}
    \item First, we provide the most comprehensive evaluation to date of causal, open‑access LLMs for mode choice prediction. We benchmark eleven open-access models (1–12B parameters) across three stated and revealed preference datasets (two open-access and one closed-access), testing 396 model configurations and generating over 79,000 mode choice predictions.
    \item Second, we introduce \textit{LiTransMC}, to the best of our knowledge, the first fine‑tuned causal LLM specialized for mode choice classification. By applying parameter‑efficient low‑rank adaptation (LoRA) with loss masking, we adapt a 12B parameter model using a modest corpus of training examples, achieving superior weighted F1 scores and distributional calibration compared with untuned local models and proprietary GPT‑4o alternatives.
    \item Third, beyond predictive accuracy, we develop a systematic framework for analyzing LLM reasoning. Using BERTopic and a new Explanation Strength Index, we quantify how LLMs articulate decision factors and align them with behavioural theory.
\end{itemize}
    
The rest of the paper is organized as follows. Section \ref{sec:Methods} details the methodological framework, including the experimental design, dataset selection, learning strategies, and local LLM setup. Section \ref{sec:Results} presents and discusses the results, focusing on predictive accuracy, reasoning performance, and the effects of fine-tuning. Section \ref{sec: recomendations} outlines key recommendations and implications for transport behaviour research and practice, and Section \ref{sec:conclusions} concludes the paper with final reflections and directions for future work.

\section{Methods}
\label{sec:Methods}
Here we present the overall experimental design, localized LLM and server setup, evaluation framework, and LLM model finetuning process.

\subsection{Experiment Design}
\label{subsection: Experiment Design}

\begin{figure}[!ht] 
    \centering
    \includegraphics[width={1\linewidth}]{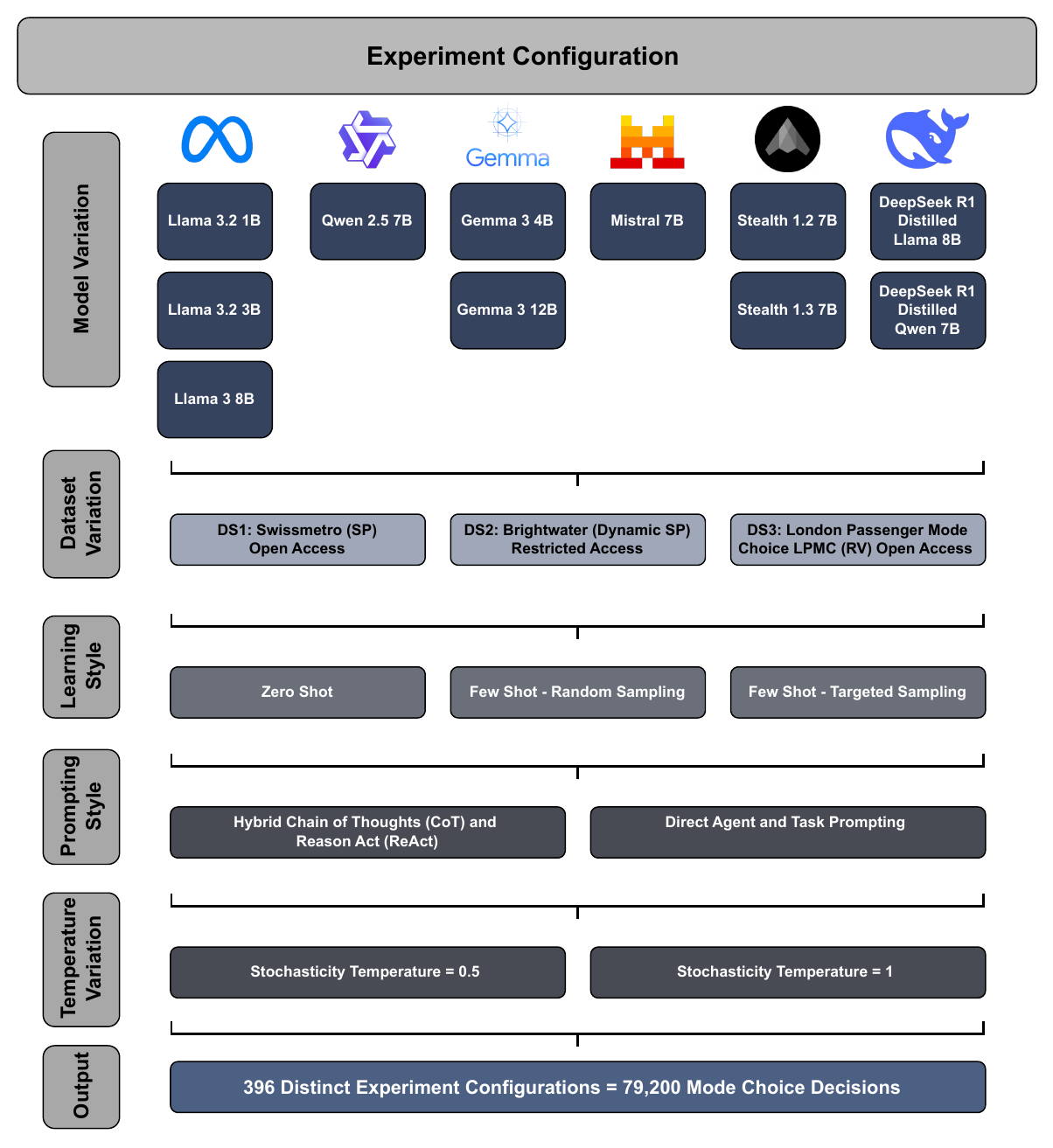}
    \caption{Overall Experiment Design}
    \label{fig: Experiment Design}
\end{figure}

To evaluate the ability of causal LLMs to model the complex relationship between commuters sociodemographics, trip attributes, and mode choice, and to examine the factors that influence the model's predictive and reasoning powers, the experiment was developed with a wide range of variations. These included varying the chosen foundational models, the size of the model parameters, training and testing datasets, learning strategies, prompting styles, and the model's temperature configuration. Figure \ref{fig: Experiment Design} outlines the overall experiment configuration and the variation in the experiment variables.

\subsubsection{Foundational Models}
Eleven foundational models from six different providers were selected, representing the most prominent small- to mid-scale open-access models. The models were chosen so that they can be efficiently deployed on high-end, consumer-level local desktop machines. Details of the selected models are provided in Table~\ref{Table:1}.
\newcolumntype{C}[1]{>{\centering\arraybackslash}m{#1}}

\begin{table}[ht]
\centering
\footnotesize
\setlength{\tabcolsep}{4pt}
\renewcommand{\arraystretch}{1.4}
\begin{tabular}{C{2.2cm} C{2.4cm} C{2.1cm} C{4.5cm} C{3.3cm}}
    \hline
    \textbf{Architecture} & \textbf{Parameters (Billion)} & \textbf{Quantization} & \textbf{Model Name} & \textbf{Reference} \\
    \hline
    LLaMA & 8 & Q4\_K\_M & Meta-Llama-3-8B-Instruct & \cite{bartowski2024llama3gguf} \\

    LLaMA & 3 & Q8\_0 & Meta-Llama-3.2-3B-Instruct & \cite{bartowski2024llama32gguf} \\  

    LLaMA & 1 & Q8\_0 & Meta-Llama-3.2-1B-Instruct & \cite{bartowski2024llama321gguf} \\

    Qwen2 & 7 & Q4\_K\_M & Qwen2.5-7B-Instruct & \cite{bartowski2024qwen25_7b_gguf} \\

    Gemma3 & 4 & Q4\_K\_M & gemma-3-4b-it & \cite{lmstudio2024gemma34bit} \\

    Gemma3 & 12 & Q3\_K\_L & gemma-3-12b-it & \cite{lmstudio2024gemma312bit} \\

    LlaMA & 7 & Q4\_K\_M & Mistral-7B-Instruct-v0.3 & \cite{bartowski2024mistral7b} \\

    LlaMA & 7 & Q4\_K\_M & stealth-v1.2 & \cite{janhq2024stealth12} \\

    LlaMA & 7 & Q8\_0 & stealth-v1.3.Q8\_0 & \cite{janhq2024stealth13} \\

    LlaMA & 8 & Q4\_K\_M & DeepSeek-R1-Distill-Llama-8B & \cite{lmstudio2024deepseek8b} \\

    Qwen2 & 7 & Q4\_K\_M & DeepSeek-R1-Distill-Qwen-7B & \cite{lmstudio2024deepseekqwen7b} \\
    \hline
\end{tabular}
\vspace{1em}
\caption{ Selected Causal LLMs for the Mode Choice Prediction Experiment}
\label{Table:1}
\end{table}

\subsubsection{Datasets}
Three distinct datasets were selected to evaluate the ability of causal LLMs to predict commuters mode choice and generate corresponding reasoning. The selection aimed to ensure diversity in both survey types and data accessibility, covering stated preference (SP) and revealed preference (RP) surveys, with open and closed data sources. This structure was intended to assess how variations in survey methodology and data availability might influence model performance, including the potential effect of prior model exposure to publicly available datasets.

To represent the open-access SP dataset, the widely used Swissmetro survey was selected \citep{bierlaire2001swissmetro}. For the closed-access SP dataset, a recent experiment conducted in the Brightwater community was utilized. Brightwater is a master-planned, mixed-use development spanning 72 acres of waterfront land in Port Credit, Mississauga, Ontario, Canada. The community is expected to be completed by 2029 and will include over 2,900 residential units and approximately 300,000 square feet of commercial space, with the first residents occupying the space in 2023.
The associated data collection campaign for this study was conducted from October 31 to November 20, 2022. A total of 159 future residents completed the survey, which gathered key sociodemographic information, attitudinal and perception indicators, as well as scenario-based travel responses from Brightwater to major neighbouring destinations across varying seasons and weather conditions.

To represent open-access RP data, the London Passenger Mode Choice (LPMC) dataset was selected \citep{hillel2018modechoice}. This dataset was chosen due to its broad application in mode choice research and its distinctive structure. Unlike most RP surveys, the LPMC dataset includes both the observed choices and the corresponding availability sets of travel alternatives, allowing to construct a situational choice sets.

For each dataset, a sample of all scenarios or trips for a 100 respondents was selected to construct the training dataset. This training set served as the source pool for generating few-shot examples used in the LLM prompting process. An additional random sample of 200 observations was drawn from the remaining data and used as the testing dataset. This approach ensures that the testing sample size remains consistent across all datasets, while the total size of the training dataset varies depending on the number of scenarios or trips reported per respondent in each survey for a 100 respondent profiles. 

\subsubsection{Learning Approach}
To evaluate the most effective learning approach, three different shot types were tested: \emph{zero-shot, few-shot with random sampling, and few-shot with targeted, similarity based sampling}. In the zero-shot experiments, the commuter’s sociodemographic profile, situational trip attributes, and a contextual description of the travel scenario were provided to the LLM, which acted as a synthetic commuter tasked with selecting the most appropriate mode of transport for the given context. In this setup, the model receives no examples from the original survey, relying solely on its internal knowledge to infer the relationship between individual characteristics, trip attributes, and mode choice. Thus, it makes decisions based entirely on the presented context and its generalized understanding, without exposure to any observed examples of human decision making from the training data. On the other hand, the few-shot approach involves providing selected examples from the training dataset to the LLM in order to support its understanding of the relationships between sociodemographic profiles, trip attributes, and mode choice decisions. Within this few-shot learning modality, two sampling strategies were employed to select examples from the training pool.

The first method involves a random selection of five examples for each synthetic commuter, providing the model with broad exposure to diverse mode choice patterns. The second method, referred to as targeted similarity-based sampling, selects examples that are most similar to the test case based on key sociodemographic and trip-related attributes. This approach is designed to guide the model with contextually relevant patterns, thereby enhancing its ability to generalize mode choice behaviour in scenarios that closely resemble the presented commuter profile and travel context.

To implement the targeted similarity-based few-shot learning approach, a multi dimensional scoring framework was developed to identify training examples most relevant to each test instance. \( d_i \) and every candidate training instance \( d_j \) is represented by a combination of sociodemographic, trip, and contextual variables. The overall similarity \( \text{Sim}(d_i, d_j) \) is calculated as a weighted sum based on a set of components:

\begin{itemize}
    \item \( S_{\text{socio}}(d_i, d_j) \): similarity based on sociodemographic variables,
    \item \( S_{\text{trip\_num}}(d_i, d_j) \): similarity based on numeric trip variables,
    \item \( S_{\text{trip\_cat}}(d_i, d_j) \): similarity based on categorical trip variables,
    \item \( S_{\text{add}}(d_i, d_j) \): similarity based on additional travel-related variables.
\end{itemize}

Each component is weighted to produce the total similarity:

\begin{equation}
\text{Sim}(d_i, d_j) = w_{\text{socio}} \cdot S_{\text{socio}}(d_i, d_j) + 
                       w_{\text{trip\_num}} \cdot S_{\text{trip\_num}}(d_i, d_j) +
                       w_{\text{trip\_cat}} \cdot S_{\text{trip\_cat}}(d_i, d_j) + 
                       w_{\text{add}} \cdot S_{\text{add}}(d_i, d_j)
                       \label{eq:gensimilarity}
\end{equation}

\noindent For example, in the Swissmetro dataset, the weights are defined as:

\[
w_{\text{socio}} = 0.35, \quad
w_{\text{trip\_num}} = 0.30, \quad
w_{\text{trip\_cat}} = 0.15, \quad
w_{\text{add}} = 0.20
\]

These weights are normalized to sum to 1.0, ensuring \( \text{Sim}(d_i, d_j) \in [0, 1] \).

For ordinal attributes (e.g., age or income), the similarity is based on discrete proximity:

\begin{equation}
S_{\text{ordinal}}(v_i, v_j) =
\begin{cases}
1.0 & \text{if } |v_i - v_j| = 0 \\
0.5 & \text{if } |v_i - v_j| = 1 \\
0.0 & \text{otherwise}
\end{cases}
\label{eq:ordinal}
\end{equation}

For continuous numeric trip variables (e.g., travel time or cost), similarity is based on inverse Euclidean distance over Min-Max normalized vectors \( \mathbf{x}_i \) and \( \mathbf{x}_j \):

\begin{equation}
S_{\text{trip\_num}}(d_i, d_j) = \frac{1}{1 + \lVert \mathbf{x}_i - \mathbf{x}_j \rVert}
\label{eq:similarity}
\end{equation}

\noindent where \( \mathbf{x}_i \) and \( \mathbf{x}_j \) represent the normalized feature vectors of the numeric attributes for \( d_i \) and \( d_j \), respectively.

The top \(k=5 \) training instances with the highest similarity scores are selected as few-shot examples and included in the prompt presented to the LLM. These examples provide context that helps the model reason about mode choice behaviour under conditions similar to those faced by the test instance.
Importantly, the exact set of categories and variables included in each similarity component may vary across datasets, depending on the availability and richness of attributes collected in each travel survey. The framework adapts to these differences by adjusting the variable groups used in similarity computation, while preserving the categorization and classification logic.

Before finalizing the current similarity framework, several alternative approaches were explored, including cosine similarity and cross-encoder embeddings for full-text comparison. While cosine similarity is widely used in natural language processing for comparing semantic vectors, and in previous studies to select the few-shot examples in mode choice experiments \citep{liu2025aligning}, it underperforms when applied to numeric survey responses, such as travel time or cost. This is because cosine similarity measures the angle between vectors rather than their magnitude, making it insensitive to absolute differences that are critical for behavioural decisions in mode choice modelling.

For instance, if we consider a test instance with a scaled travel cost of 0.81 (representing 240 CHF). Two candidate training examples have scaled values of 0.8 (237 CHF) and 0.20 (65 CHF), respectively. Under cosine similarity, both vectors yield high similarity scores (e.g., \( > 0.98 \)) because their directions are aligned, despite the second case representing a cost difference of 175 CHF. In contrast, using inverse Euclidean distance, the similarity scores are more behaviourally meaningful: 
\[
\text{Sim}_{\text{inv-Euc}}(240, 237) = \frac{1}{1 + 0.01} = 0.99, \quad \text{Sim}_{\text{inv-Euc}}(240, 65) = \frac{1}{1 + 0.60} \approx 0.625
\]
This approach correctly penalizes large differences in magnitude, which are essential for representing cost and time sensitivity in human decision-making.

For categorical and ordinal variables, cosine similarity similarly fails to capture behavioural meaning, as it treats differences between categories like ``leisure'' and ``business'', or between income levels 2 and 3, as abstract angular dissimilarities rather than ordered or discrete groupings. The adopted framework, on the other hand, uses exact matching for nominal variables and ordinal proximity scoring for ordered variables as described previously in Equation~\ref{eq:ordinal}, thereby better reflecting real-world perception thresholds (e.g., adjacent age or income groups are more similar than distant ones).

Another element to consider lies in the design of the prompt construction pipeline. Under zero-shot and random few-shot configurations, survey responses are pre-processed by converting structured attributes (both categorical and continuous) into full textual descriptions. These narrative prompts are stored in a textual database and retrieved directly for LLM calls, minimizing computational cost during runtime. However, for the targeted similarity-based setup, the similarity computation is performed on the original structured data (prior to textual transformation). For each test instance, the top-\(k\) most similar training samples are first identified, and their corresponding prompts are dynamically constructed at runtime. This ensures that the similarity is calculated without semantic noise from full-text embeddings.
\subsubsection{Prompting Style}
To evaluate the effect of prompting style on LLM's predictive performance, two distinct styles were implemented. The first adopts a hybrid structure inspired by Chain-of-Thought (CoT) and the Reason+Act (ReACT) prompting, wherein the model is encouraged to simulate a deliberative reasoning process before arriving at a decision. In this format, the prompt explicitly instructs the LLM to consider relevant trade-offs and contextual factors, such as time, cost, and purpose, based on the travel scenario and commuter profile, and to articulate a rationale before stating its selected mode. This approach facilitates cognitive transparency by allowing the model to express its internal reasoning prior to outputting a final decision.

The second prompting style adopts a direct decision format, wherein the LLM is presented with the full scenario description including commuter attributes and available alternatives, and is asked to output only the selected mode, without generating intermediate reasoning. This method prioritizes succinctness and emulates standard classification behaviour, allowing for rapid assessments of model preference without the cognitive load in the process. Worth noting that in both styles, the models are still instructed to output a separate JSON briefly outlining the rationale behind their selected choice. The difference here lies in the intermediary process, where the first style requires the model to follow a multi-step reasoning prior their answer, while the second uses more direct approach. 

\subsubsection{Temperature Settings}
In addition to prompt style, the temperature parameter of the LLM was varied to explore the influence of generative randomness on prediction outcomes. In this context, temperature controls the stochasticity of the model’s token sampling process, with lower values (e.g., 0.5) promoting more deterministic with focused outputs, and higher values (e.g., 1.0) allowing for greater variation and exploratory generation. Temperature $0.5$ was selected to assess performance under conservative, reproducible conditions, while temperature $1.0$ was included to evaluate whether increased generative diversity may enhance generalization, particularly in cases involving ambiguity or competing trade-offs in the mode choice scenario.

Combining all experimental dimensions results in a comprehensive evaluation matrix. A total of eleven foundational LLMs were tested across three datasets (Swissmetro, Brightwater SP, and LPMC RP), each under three learning modalities: zero-shot, random few-shot, and targeted few-shot. For each learning style, two prompting strategies (CoT-ReACT hybrid and direct) and two temperature settings (0.5 and 1.0) were applied, resulting in:

\begin{align*}
11 \, \text{models} \times 3 \, \text{datasets} &\times 3 \, \text{learning styles} \times 2 \, \text{prompting styles} \times 2 \, \text{temperatures} \\
&= 396 \, 
\text{unique experiments}
\end{align*}

Under each experiment, the model was tasked with 200 prediction and reasoning calls using the designated test set, yielding a total of:

\[
396 \, \text{configurations} \times 200 \, \text{test instances} = 79{,}200 \, \text{ Mode choice decisions}
\]

Each experiment was subsequently evaluated by comparing the model generated predictions against the actual survey respondents’ mode choices. This evaluation framework allows for a structured and systematic assessment of each LLM's predictive performance under varying learning strategies, prompting styles, and temperature settings. This allows the investigation into the behavioural fidelity of LLMs in emulating human travel decision making and highlights the conditions under which these models can offer the most accurate and generalizable representations of mode choice behaviour. 

\subsection{Local LLMs  Setup and Inference Framework}

To simulate mode choice decisions using locally hosted LLMs, synthetic LLM commuting agents are generated by embedding real survey responses into structured prompts. The objective is to generate $N$ mode choice predictions for each experimental configuration. Below are the definitions of the parameters involved:

\begin{itemize}
    \item $\mathbf{M}$: a locally hosted open-access LLM.
    \item $d_i$: structured data of agent $i$, including sociodemographic and trip attributes drawn from the original survey dataset.
    \item $S$: the system-level prompt that defines the LLM's expected role and behaviour.
    \item $U(d_i)$: a transformation function that converts $d_i$ into natural language format for prompt construction.
    \item $E_i$: a set of $k$ few-shot examples included in the prompt for agent $i$.
    \item $Q_i$: the complete prompt for agent $i$, composed as $Q_i = (S, U(d_i))$ in the zero-shot case or $Q_i = (S, U(E_i), U(d_i))$ in the few-shot case.
    \item $f(\mathbf{M}, Q_i)$: function that queries model $\mathbf{M}$ with $Q_i$ and returns the raw output text $R_i$.
    \item $g(R_i)$: function that parses raw text $R_i$ into structured JSON $J_i$.
    \item $\sigma(J_i)$: function that extracts the predicted travel mode $\hat{y}_i$ from $J_i$.
    \item $\rho(J_i)$: function that extracts the LLM's explanation or rationale $z_i$ for the selected mode from $J_i$.
\end{itemize}

\subsubsection*{Zero-Shot Learning}

In zero-shot inference, the model receives no prior examples. The inference process proceeds as follows:

\begin{align}
Q_i &= (S, U(d_i)) && \text{Prompt construction} \\
R_i &= f(\mathbf{M}, Q_i) && \text{Model call} \\
J_i &= g(R_i) && \text{Parse model response} \\
\hat{y}_i &= \sigma(J_i) && \text{Extract predicted mode} \\
z_i &= \rho(J_i) && \text{Extract decision reasoning}
\end{align}

The resulting decision instance for agent $i$ is stored as:

\begin{equation}
\mathbf{r}_i = (i, d_i, \hat{y}_i, z_i)
\end{equation}

All $N$ predictions form:

\begin{equation}
\text{Database} =
\begin{bmatrix}
\mathbf{r}_1 \\
\mathbf{r}_2 \\
\vdots \\
\mathbf{r}_N
\end{bmatrix}
\end{equation}

\subsubsection*{Few-Shot Learning}

In few-shot prompting, a set of $k$ examples $E_i$ is provided alongside each agent's prompt. Two methods are used to select $E_i$:

\paragraph{1. Random Sampling}

A set $E_i^{\text{rand}}$ is drawn uniformly at random from the training pool $\mathcal{T}$:

\begin{equation}
E_i^{\text{rand}} = \text{RandomSample}(\mathcal{T}, k)
\end{equation}

\paragraph{2. Targeted Similarity-Based Sampling}

This method selects the top-$k$ training samples most similar to agent $i$ based on structured data $d_i$. As defined in Section \ref{subsection: Experiment Design}, Equation \ref{eq:gensimilarity}. 

The top-$k$ training examples with the highest similarity scores are selected as:

\begin{equation}
E_i^{\text{sim}} = \text{TopK}_{d_j \in \mathcal{T}}(\text{Sim}(d_i, d_j))
\end{equation}

The resulting prompt for few-shot becomes:

\begin{equation}
Q_i = (S, U(E_i), U(d_i)), \quad \text{where } E_i \in \{E_i^{\text{rand}}, E_i^{\text{sim}}\}
\end{equation}

\vspace{0.5em}

\begin{figure}[!ht] 
    \centering
    \includegraphics[width={1\linewidth}]{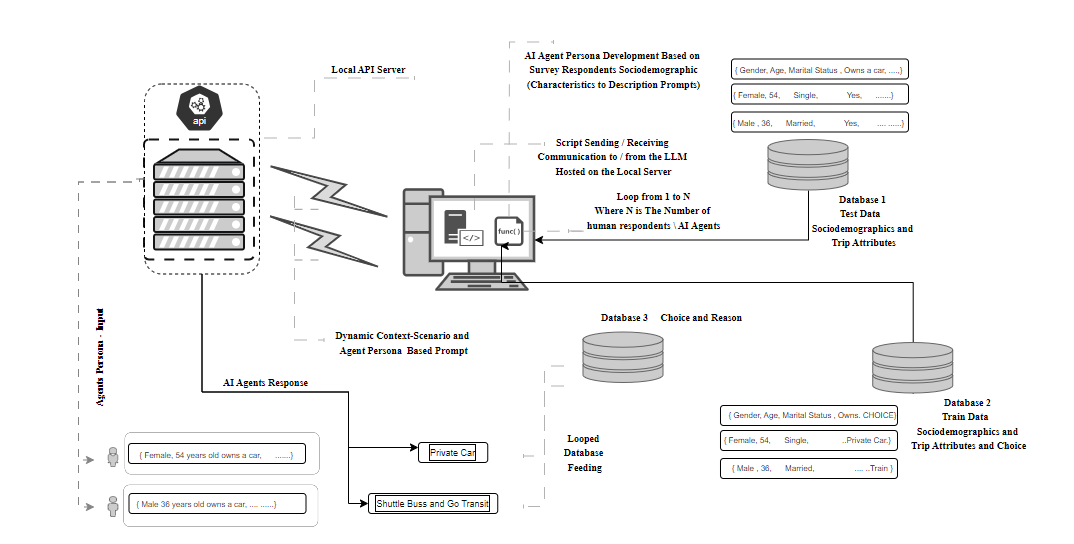}
    \caption{Localized LLM and Server Setup}
    \label{fig: Localized LLM and Server Setup}
\end{figure}

Figure~\ref{fig: Localized LLM and Server Setup} illustrates the full local inference setup used in this study. It presents the pipeline from structured survey data ingestion to LLM querying and storage of predictions. 

\subsection{ Performance Evaluation Framework}
\label{sec:eval-prediction}
In this framework, LLMs are evaluated for both prediction and reasoning. The evaluation comprises two complementary components: \textit{(i)} predictive performance assessment, which measures accuracy and distributional alignment with observed mode choices (Section~\ref{subsec:pred-perf}), and \textit{(ii)} reasoning performance assessment, which examines the structure, strength, and thematic coherence of the model-generated explanations (Section~\ref{sec:reasoning-bertopic}).

\subsubsection{LLMs Predictive Performance Evaluation}
\label{subsec:pred-perf}
To assess the LLM's predictive performance for travel mode choice preferences, we implemented a two-tiered evaluation framework. This framework considers both instance level accuracy and the overall distributional fidelity of predicted mode shares. In addition to predictions, the framework is also designed to handle model generated reasoning,

Let \( \mathcal{D}_{\text{test}} = \{(d_i, y_i)\}_{i=1}^N \) be the test dataset, where \( d_i \) is the input data (trip and sociodemographic attributes), and \( y_i \in \{1, 2, \dots, C\} \) denotes the ground-truth mode label and \( C \) denote the total number of transport mode categories used in the classification task. Each evaluated model \( \mathcal{M}^{(m)} \) generates a predicted mode \( \hat{y}_i^{(m)} \) for each instance \( i \). We compute the following standard classification metrics:

\begin{equation}
\text{Accuracy}^{(m)} = \frac{1}{N} \sum_{i=1}^N \mathbf{1}(y_i = \hat{y}_i^{(m)})
\end{equation}

\begin{equation}
\text{Precision}^{(m)} = \frac{1}{C} \sum_{c=1}^C \frac{\text{TP}_c^{(m)}}{\text{TP}_c^{(m)} + \text{FP}_c^{(m)}}
\end{equation}
Here, \( \text{TP}_c^{(m)} \) is the number of true positives, and \( \text{FP}_c^{(m)} \) is the number of false positives for class \( c \).

\begin{equation}
\text{Recall}^{(m)} = \frac{1}{C} \sum_{c=1}^C \frac{\text{TP}_c^{(m)}}{\text{TP}_c^{(m)} + \text{FN}_c^{(m)}}
\end{equation}
\( \text{FN}_c^{(m)} \) denotes the number of false negatives for class \( c \).

\begin{equation}
\text{F1}^{(m)} = \frac{1}{C} \sum_{c=1}^C \frac{2 \cdot \text{Precision}_c^{(m)} \cdot \text{Recall}_c^{(m)}}{\text{Precision}_c^{(m)} + \text{Recall}_c^{(m)}}
\end{equation}

\begin{equation}
\text{F1}^{(m)}_{\text{weighted}} = \sum_{c=1}^C w_c \cdot \text{F1}_c^{(m)}
\end{equation}

The weighted F1 Score accounts for class imbalance by weighting each class-specific F1 Score by its support \( w_c = \frac{n_c}{N} \), where \( n_c \) is the number of ground truth instances in class \( c \).
In addition to these instance-level metrics, we compute distribution-level divergence between the predicted and true mode shares. 
Let the empirical ground-truth distribution over classes be defined as
\begin{equation}
    p_c = \frac{1}{N} \sum_{i=1}^N \mathbf{1}(y_i = c),
\end{equation}
and the predicted distribution by model \( \mathcal{M}^{(m)} \) as
\begin{equation}
    \hat{p}_c^{(m)} = \frac{1}{N} \sum_{i=1}^N \mathbf{1}(\hat{y}_i^{(m)} = c).
\end{equation}

To address zero-count issues in the computation of divergence-based metrics, we apply Laplace smoothing using a small constant \( \varepsilon = 10^{-9} \), yielding:
\begin{equation}
    p_c^{\text{smooth}} = \frac{p_c + \varepsilon}{1 + 3\varepsilon}, \quad 
    \hat{p}_c^{(m), \text{smooth}} = \frac{\hat{p}_c^{(m)} + \varepsilon}{1 + 3\varepsilon}.
\end{equation}

We then compute the following distribution-level evaluation metrics:
\begin{equation}
    \text{DistMAE}^{(m)} = \frac{1}{C} \sum_{c=1}^C \left| p_c - \hat{p}_c^{(m)} \right|.
\end{equation}

\begin{equation}
    \text{JSD}^{(m)} = \frac{1}{2} \left[ 
        \sum_{c=1}^C p_c^{\text{smooth}} \log \left( \frac{p_c^{\text{smooth}}}{M_c} \right) +
        \hat{p}_c^{(m), \text{smooth}} \log \left( \frac{\hat{p}_c^{(m), \text{smooth}}}{M_c} \right)
    \right],
\end{equation}
where \( M_c = \frac{1}{2} \left( p_c^{\text{smooth}} + \hat{p}_c^{(m), \text{smooth}} \right) \) is the average distribution.

\begin{equation}
    \text{CE}^{(m)} = - \sum_{c=1}^C p_c^{\text{smooth}} \log \left( \hat{p}_c^{(m), \text{smooth}} \right).
\end{equation}

 Each metric in the evaluation framework was selected for its ability to capture distinct yet complementary aspects of model performance. The accuracy metric. provides an overall correctness measure but can be biased by class imbalance. The precision and recall help evaluate model performance on individual classes, with F1\_Macro offering a class-averaged harmonic mean and F1\_Weighted accounting for class imbalance by weighting each class's F1 score by its support. On the distribution level, MAE quantifies the average deviation between predicted and true mode share proportions, while cross-entropy penalizes probabilistic mismatch and is sensitive to over or under-confidence in predictions. JSD, on the other hand a symmetrized and smoothed variant of Kullback-Leibler divergence, offers a bounded and interpretable measure of similarity between the predicted and actual distributions \citep{lin2002divergence}.

\subsubsection{LLMs Reasoning Performance Evaluation}
\label{sec:reasoning-bertopic}

To complement the evaluation of predicted travel modes, we conducted a detailed reasoning analysis to assess the structure, diversity, and strength of the natural language justifications generated by each model. This analysis consisted of two main components: (i) computation of an explanation strength index, and (ii) topic modelling using BERTopic \citep{grootendorst2022bertopic}.

For each generated reasoning \( z_i \) from instance \( i \), we defined an Explanation Strength Index \( \mathrm{ESI}(z_i) \) that quantifies the extent to which the explanation references utility-relevant decision factors. Let \( \mathcal{F} = \{\text{time}, \text{cost}, \text{comfort}, \text{convenience},\) \(\text{frequency} \} \) denote the set of predefined decision-related keywords. The \( \mathrm{ESI}(z_i) \) is computed as:

\begin{equation}
\mathrm{ESI}(z_i) = \frac{1}{|\mathcal{F}|} \sum_{f \in \mathcal{F}} \mathbb{1}_{\{f \in z_i\}},
\end{equation}

where \( \mathbb{1}_{\{f \in z_i\}} \) is an indicator function equal to 1 if factor \( f \) appears in \( z_i \), and 0 otherwise. The ESI score lies between 0 and 1, with higher values indicating a denser reference to decision-relevant concepts.

Additionally, each reasoning \( z_i \) was encoded into a high-dimensional semantic vector \( \mathbf{e}_i \in \mathbb{R}^{d} \) using the \texttt{all-MiniLM-L6-v2} sentence transformer model. These embeddings were reduced to two dimensions via Uniform Manifold Approximation and Projection (UMAP) for visualization:

\begin{equation}
\mathbf{u}_i = \mathrm{UMAP}(\mathbf{e}_i) \in \mathbb{R}^2,
\end{equation}

with UMAP parameters set to \( n_{\text{neighbors}} = 15 \), \( \text{min\_dist} = 0.1 \), and cosine distance as the metric.

The BERTopic algorithm was then applied using the sentence embeddings and UMAP-reduced vectors. This model clustered the explanations into coherent topics \( T = \{t_1, t_2, \dots, t_K\} \), with each instance assigned a topic label \( t_i \in T \). For each topic, a ranked list of representative keywords was extracted using class-based TF-IDF, for enhanced interpretability.

These different evaluation dimensions enabled a structured evaluation of the interpretive dimension of model outputs, which allowed us to comprehensively assess  the content richness and diversity of the generated rationales along with the models predictive accuracy.

\subsection{Fine-tunning}
The foundation for this research is the decoder-only transformer LLMs a.k.a. Casual LLMs. Its extensive pre-training provides a rich basis of general knowledge, which we specialize for travel mode choice prediction task through finetuning.

The training corpus consists of several thousand structured examples derived from the Swissmetro travel survey responses except for a hold sample that was used for testing. Each example is composed of: 1) an instruction, a detailed text prompt describing a traveller's sociodemographic profile and a specific travel scenario with the transport options and their attributes (e.g., travel time, cost, frequency); and 2) a selected\_mode, the ground-truth label corresponding to the mode choice.

For robust model development, the dataset was partitioned into a training set (90\%) and a validation set (10\%). This allows for performance monitoring on unseen data to prevent overfitting and to select the best-performing model checkpoint. This is separate from the 200 test set that was set aside for the finetuned model validation and comparison, similar to the process described in Section \ref{sec:eval-prediction}.

\subsubsection{Parameter Efficient Finetuning (PEFT) Strategy: QLoRA}

Our parameter-efficient finetuning strategy is based on Quantized Low-Rank Adaptation (QLoRA) \citep{dettmers2023qlora}, an advanced method that enhances the original Low-Rank Adaptation (LoRA) framework \citep{hu2022lora} by integrating 4-bit quantization of the base model's weights, thereby significantly reducing memory requirements. This strategy makes finetuning large models feasible on limited hardware (a single GPU with 12GB VRAM used for this research) by drastically reducing the number of trainable parameters.

The primary memory bottleneck is the storage of the base model's weights. QLoRA addresses this by loading the pre-trained model with its weights quantized to a 4-bit precision using the NormalFloat4 (NF4) data type. This reduces the memory footprint by a factor of four compared to 16-bit precision. During computation, these 4-bit weights are de-quantized on-the-fly to a higher precision compute data type (float16) to maintain numerical stability during the forward and backward passes.

More broadly, LoRA operates on the principle that the change in model weights during adaptation, denoted as \( \Delta W \), has a low intrinsic rank \citep{hu2022lora}. Instead of training the full weight matrix \( W_0 \in \mathbb{R}^{m \times n} \), LoRA freezes \( W_0 \) and injects smaller, trainable adapter matrices into the model's layers. The weight update is represented by a low-rank decomposition:
\begin{equation}
\Delta W = V \cdot U
\end{equation}
where \( U \in \mathbb{R}^{r \times n} \) and \( V \in \mathbb{R}^{m \times r} \) are the trainable low-rank matrices, and the rank \( r \ll \min(m, n) \). The modified forward pass for a given layer includes the scaled LoRA update:
\begin{equation}
h = W_0 z + \frac{\alpha}{r} V U z
\end{equation}
For this study, we configured LoRA with a rank \( r = 32 \) and a scaling factor \( \alpha = 64 \). These trainable matrices were applied to all major linear layers of the model architecture, including query, key, value, and output projections, as well as the feed-forward network layers.

\subsubsection{Training Objective and Label Leakage Mitigation}
As the model was finetuned using a Causal LLM objective, a significant challenge when applying this objective to structured prediction tasks with decoder-only architectures is label leakage \citep{raffel2020exploring, li2023label}. This occurs when the ground-truth answer is exposed to the model as part of its input context, allowing the model to achieve high performance by simply copying the answer rather than learning the task.

To resolve this, we implemented a loss masking strategy. While the model receives the full sequence \([ \text{Prompt}, \text{Answer} ]\) as input to maintain conversational context, the loss function is computed only for the Answer portion. Let a sequence of tokens be \( \boldsymbol{\tau} = (\tau_1, \tau_2, \dots, \tau_{|Q|}) \) and its corresponding labels be \( \boldsymbol{\lambda} = (\lambda_1, \lambda_2, \dots, \lambda_{|Q|}) \), where the labels for prompt tokens are set to an ignore index of \(-100\). The masked cross-entropy loss is formulated as:
\begin{equation}
\mathcal{L}_{\text{masked}}(\theta) = - \sum_{j=1}^{|Q|} \mathbb{I}(\lambda_j \neq -100) \cdot \log P(\tau_j \mid \tau_{<j}; \theta)
\end{equation}
where \( \theta \) represents the model's trainable parameters, \( P(\tau_j \mid \tau_{<j}; \theta) \) is the predicted probability of token \( \tau_j \), and \( \mathbb{I}(\cdot) \) is an indicator function that equals 1 if the condition is true and 0 otherwise. This ensures that the model parameters are updated exclusively based on their ability to correctly predict the answer tokens.

As for the training configuration, the model was trained for a maximum of 5 epochs using the Paged AdamW optimizer, a memory-efficient variant introduced in the QLoRA framework \citep{dettmers2023qlora}, which builds upon the AdamW algorithm with decoupled weight decay \citep{loshchilov2017decoupled}. A constant learning rate of \( 2 \times 10^{-5} \) was used.
Performance was evaluated on the held-out validation set at the end of each epoch using the weighted F1 score as the primary metric for model selection. An early stopping mechanism with a patience of 2 was implemented to halt training if the validation F1 score did not improve for two consecutive epochs to prevent overfitting and to select the model checkpoint that best generalizes to unseen data. Worth noting, that the reported F1 and JSD scores for the fine-tuned model in section \ref{fine-tune data} are based on the model deployment on the test data, which were not part of the training or the validation data used at the fine-tuning stage. 

\section{Results and Discussion}
\label{sec:Results}
This section presents the results of our systematic evaluation of causal LLMs in the context of transport mode choice modelling. The analysis is structured into three core components: predictive performance, reasoning quality, and finetuning effects.
First, we assess the ability of eleven open access LLMs, ranging from 1 to 12B parameters, to accurately replicate human mode choice behaviour across different datasets, learning strategies, and diverse prompting and temperature configurations. Performance is evaluated at both the instance level (e.g., accuracy, macro F1) and the distributional level (e.g., Jensen-Shannon divergence), benchmarked against observed human survey data.
Second, we analyze the quality and structure of the natural language rationales generated by each model. This includes both quantitative measures, such as the Explanation Strength Index (ESI), and semantic level insights extracted via topic modelling using BERTopic. The objective is to evaluate not only what the models predict, but why they make those predictions, and how closely those rationales align with human decision factors.
Finally, we examine the impact of fine-tuning a causal LLM on labelled travel survey data. This is the first known attempt to fine-tune a foundation model specifically for transport mode choice prediction. We compare its performance to base models and highlight the extent to which supervised adaptation improves alignment with ground truth selections.

\subsection{Instant-Level LLMs Mode Choice Predictive Performance Evaluation}
\label{sec:predictive-performance}

\subsubsection{F1 Scores and Model's Consistency}
\begin{figure}[!ht] 
    \centering
    \includegraphics[width={\linewidth}]{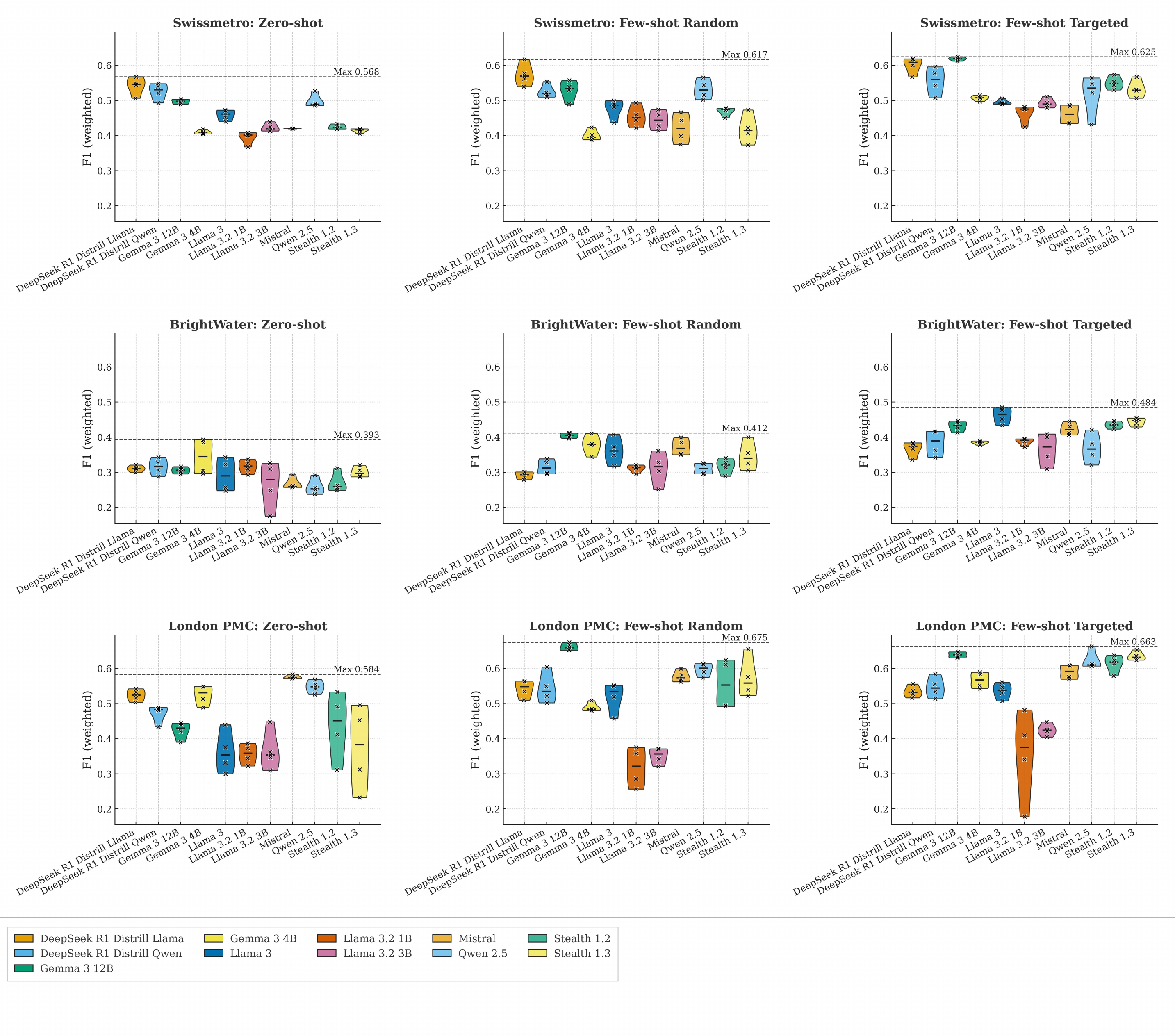}
    \caption{Instant Level LLMs Mode Choice Predictive Performance Evaluation}
    \label{fig: Instant Level LLMs Mode Choice Predictive Performance Evaluation}
\end{figure}

Figure~\ref{fig: Instant Level LLMs Mode Choice Predictive Performance Evaluation} presents the distribution of weighted F1 scores achieved by each model across three travel survey datasets; Swissmetro, Brightwater, and London PMC under three learning styles: zero-shot, random few-shot, and targeted few-shot. The violin plots show each model's distribution across four experimental runs, highlighting central tendency (mean), maximum performance (peak), and dispersion (IQR). The vertical span of each violin reflects the full value range across runs. This structure allows us to simultaneously evaluate performance levels and consistency across models and prompting styles while being able to assess the model's sensitivity to varying prompting styles and temperature. For the remainder of this section, the results are presented and discussed dataset-wise, while Table  \ref{tab:all-datasets-summary} presents the summary, mean, peak and sensitivity/ consistency for top performing models.  

For the Swissmetro dataset, zero-shot experiments, DeepSeek R1 Distill Llama 8B outperforms others with a mean of 0.542 and peak of 0.568, maintaining a narrow IQR of 0.018, which indicates robust and insensitive performance under varying prompt conditions and temperature. Gemma 3 12B follows with a mean value of 0.497 and an even lower IQR of 0.011. Meanwhile, smaller models like Llama 3.2 1B record lower means (0.395) but also low dispersion (IQR = 0.011), which suggests that their output is rigid with limited prediction capacity at the base level, especially without any guiding examples.

When transitioning to random few-shot learning, most models show gains in mean performance, though often at the cost of increased IQR or dispersion of performance results. This could mainly be attributed to the variability in the presented training profiles. DeepSeek R1 Distill Llama 8B reaches a new mean of 0.574 (peak = 0.617, IQR = 0.031), while Gemma 3 12B rises to 0.529 (peak = 0.558, IQR = 0.023), still maintaining less sensitivity to the variation of the presented training profiles, albeit higher than its baseline IQR performance.

Targeted few-shot prompts significantly elevate both accuracy and stability. Gemma 3 12B reaches the highest overall mean of 0.619 (peak = 0.625), while tightening its IQR to just 0.0035, making it the most consistent top performer. DeepSeek Llama follows with a mean = 0.601 but with a higher sensitivity to the prompt style and temperature.

As for Brightwater, this travel survey dataset exhibits generally lower performance across all learning regimes. The complexity stems from its dynamic choice structure. Although each individual sample presents only 4 alternatives, the global choice set comprises 10 distinct modes. This variability, combined with relatively homogeneous commuter profiles, renders the prediction task more challenging. In the zero-shot setting, Gemma 3 4B achieves the highest mean score (0.345) and peak performance (0.393), but comparatively has a wide IQR of 0.080, indicating moderate sensitivity to prompt and temperature variations. DeepSeek R1 Distilled models, on the other hand, deliver a lower mean (0.310) but with higher consistency across the different runs (Average IQR = 0.020),

With random few-shot learning, incremental gains are observed across the board. Gemma 3 12B attains the highest mean of 0.405 and peak of 0.412 while achieving a tight IQR of 0.010. Llama 3.2 1B shows smaller improvements (mean = 0.311, peak = 0.321) with consistent spread (IQR = 0.008 ). Under the targeted few-shot, gains are noticed again with Llama 3 8B emerging as the top-performing model, achieving the highest mean (0.462) and peak (0.484) scores. 

Models performance for London PMC shows a relatively higher baseline. In the zero-shot setup, Mistral 7B leads with a mean of 0.576 and a peak of 0.584, with IQR = 0.006. Stealth 1.3 7B performs less consistently, showing a much broader IQR = 0.171 despite competitive peaks.
Random few-shot prompts elevate Gemma 3 12B to a mean of 0.661 (peak = 0.675), with an IQR of only 0.007. Meanwhile, Stealth models appear competitive with relatively high peak values, although with much higher sensitivity to the examples and/ or the variation in the prompting styles and temperatures. 
In the targeted few-shot, Gemma 3 12B acheives the highest mean (0.639) and a peak (0.647), while Qwen 2.5 7B hits the highest peak of 0.663 (mean = 0.622, IQR = 0.018). Stealth 1.3 7B here become more consistent (IQR = 0.012), and achieves a respectable overall predictive performance, which shows that they are more adaptable when prompted with tailored examples.

\begin{table}[H]
\centering
\small
\caption{Top model performance summary per learning style across datasets}
\label{tab:all-datasets-summary}
\begin{tabular}{@{}lccc@{}}
\toprule
\textbf{Regime} & \textbf{Top Mean (Model)} & \textbf{Top Peak (Model)} & \textbf{Tightest IQR (Model)} \\
\midrule
\multicolumn{4}{@{}l}{\textbf{Swissmetro}}\\
Zero-Shot & 0.542 (DeepSeek R1) & 0.568 (DeepSeek R1) & 0.011 (Gemma 3 12B) \\
Random Few-Shot & 0.574 (DeepSeek R1) & 0.617 (DeepSeek R1) & 0.023 (Gemma 3 12B) \\
Targeted Few-Shot & 0.619 (Gemma 3 12B) & 0.625 (Gemma 3 12B) & 0.0035 (Gemma 3 12B) \\
\midrule
\multicolumn{4}{@{}l}{\textbf{Brightwater}}\\
Zero-Shot & 0.345 (Gemma 3 4B) & 0.393 (Gemma 3 4B) & 0.008 (DeepSeek R1) \\
Random Few-Shot & 0.405 (Gemma 3 12B) & 0.412 (Gemma 3 12B) & 0.010 (Gemma 3 12B) \\
Targeted Few-Shot & 0.462 (Llama 3 8B) & 0.484 (Llama 3 8B) & 0.003 (Gemma 3 4B) \\
\midrule
\multicolumn{4}{@{}l}{\textbf{London PMC}}\\
Zero-Shot & 0.576 (Mistral 7B) & 0.584 (Mistral 7B) & 0.006 (Mistral 7B) \\
Random Few-Shot & 0.661 (Gemma 3 12B) & 0.675 (Gemma 3 12B) & 0.007 (Gemma 3 12B) \\
Targeted Few-Shot & 0.639 (Gemma 3 12B) & 0.663 (Qwen 2.5 7B) & 0.012 (Stealth 1.3 7B) \\
\bottomrule
\end{tabular}
\end{table}

Across all three datasets, the transition from zero-shot to random few-shot prompting consistently yields notable improvements in mean weighted F1 scores, typically ranging from +0.03 to +0.07. However, this performance gain is frequently accompanied by an increase in interquartile range, with several models exhibiting expanded variability between runs. The pattern is mainly led by the sensitivity to the quality and coherence of uncurated in-context examples, which potentially undermines the consistency despite improvements in central tendency. 
Targeted few-shot prompting on the other hand, enhances predictive accuracy and systematically improves reliability. Selecting contextually relevant exemplars based on structured multi-layered similarity, IQR values are frequently reduced, often by 0.03 to 0.10, which highlights their stabilizing effect across models. 

\subsubsection{Variance Decomposition of Predictive Performance}
\begin{figure}[!ht] 
    \centering
    \includegraphics[width= {\linewidth}]{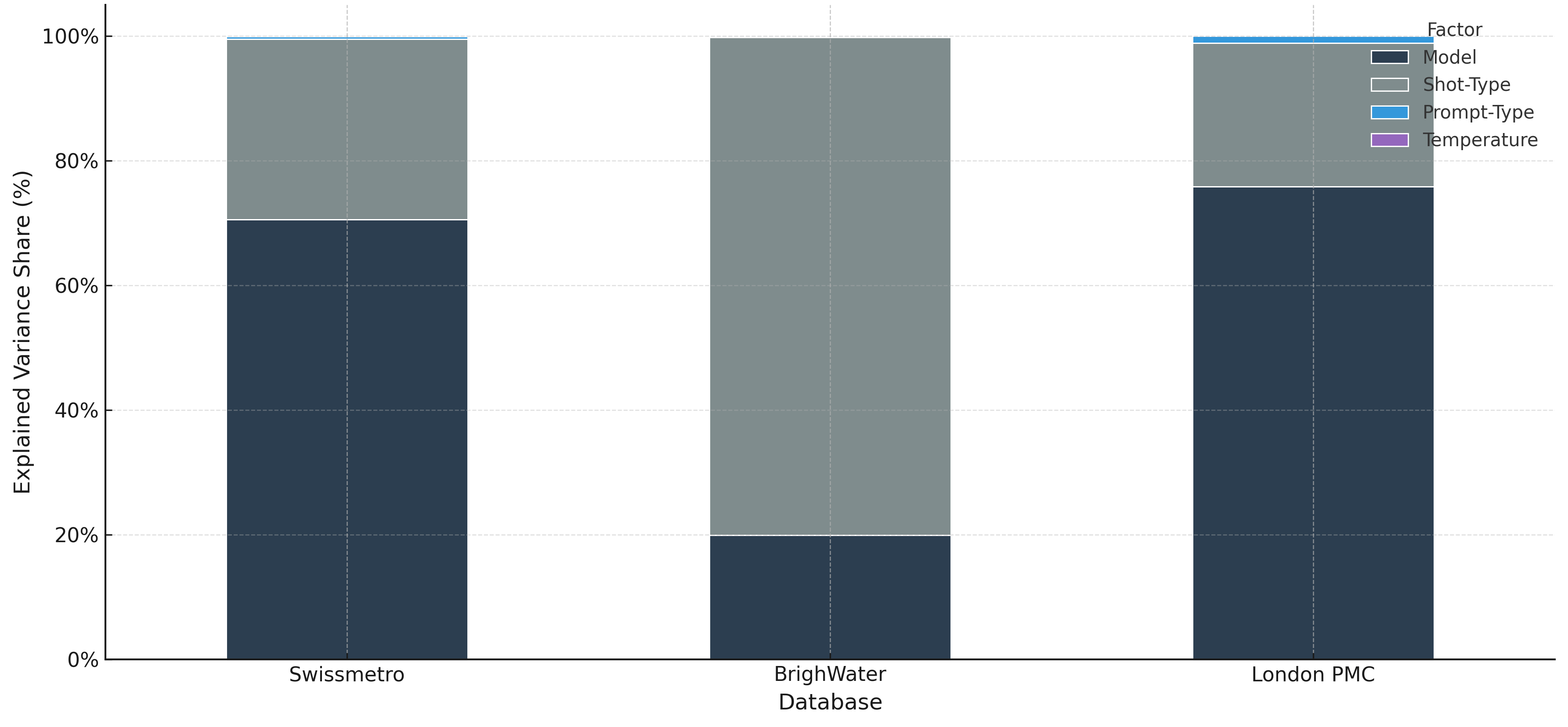}
    \caption{Decomposition of Model Performance}
    \label{fig: Variance Explanation}
\end{figure}

To further investigate the variability observed in the weighted F1 scores and identify which experimental factors most strongly influence predictive performance, we decompose the weighted F1 variance across the four key dimensions: \emph{the choice of language model (Model), the learning strategy (Shot-Type: zero-shot vs.\ few-shot), the prompt style (Prompt-style), and the model temperature (Temp)}.
Figure~\ref{fig: Variance Explanation} presents the decomposition of the variance share that can be explained by each factor across the three datasets. For each dataset, we fit an ordinary least squares (OLS) model of the form:

\begin{equation}
\text{F1}_{\text{Weighted}} \sim C(\text{Model}) + C(\text{Shot-Type}) + C(\text{Prompt-Style}) + C(\text{Temp})
\end{equation}

We then perform a Type II ANOVA to assess the relative contribution of each factor. The sum of squares attributable to each term was normalized such that the four contributions sum to 100\% of the explained variance.

In Swissmetro, the choice of model is the dominant driver of performance variability, accounting for 70.6\% of the explained variance. This indicates that the choice of a more capable model yields the greatest returns. The learning strategy or shot-type contributes 29.0\%, which constitutes a significant portion of the predictive power. Prompting style and temperatures impact compared to model choice and learning strategy are marginal, explaining just 0.36\% and 0.10\% respectively.
The analysis of the Brightwater dataset, on the other hand, exhibits a different pattern, with shot-type explaining a substantial 79.8\% of the variance. This reflects the increased difficulty of the task and the impact of in-context examples in elevating performance. Here, model choice plays a secondary role (19.9\%), while prompt-type and temperature again contribute minimally (0.21\% and 0.12\%). These findings suggest that for low-baseline tasks, few-shot examples can be transformative, whereas model choice or upgrades alone yield a less effective boost in returns.

As for the London PMC dataset, the analysis aligns more closely with Swissmetro. Model choice dominates (75.8\% of variance), followed by Shot-Type (23.0\%). Prompt-Type contributes slightly more here than in the other datasets (1.1\%), though still marginal. Temperature influence is effectively null (0.005\%). The results simply demonstrate the importance of selecting a performant LLM and a suitable learning strategy, while minor prompt adjustments or temperature tuning have limited practical value, and can be used for minor gains in terms of achieving optimal performance.

Overall, the analysis shows that for more traditional choice sets and diversified commuter profiles through larger samples, upgrading the model explains over 70\% of F1 variation as we can note in both Swissmetro and London PMC. However, for more challenging choice sets and less diverse and lower variability in commuter profiles, carefully curated few-shot examples are responsible for nearly 80\% of observed gains in the predictive performance, hence the importance of data context when model performance is initially weak. 
On the other hand, the joint contribution of prompt-type and temperature remains below 2\% in all datasets, indicating that these elements, while not irrelevant, are secondary considerations. Practitioners may use standard templates and moderate temperature settings without significant performance trade-offs. Therefore, prompt engineering and temperature tuning can be treated as last mile optimizations with marginal returns.

\subsubsection{LLM's Structure and Relative Performance Evaluation}
\begin{figure}[!ht] 
    \centering
    \includegraphics[width={\linewidth}]{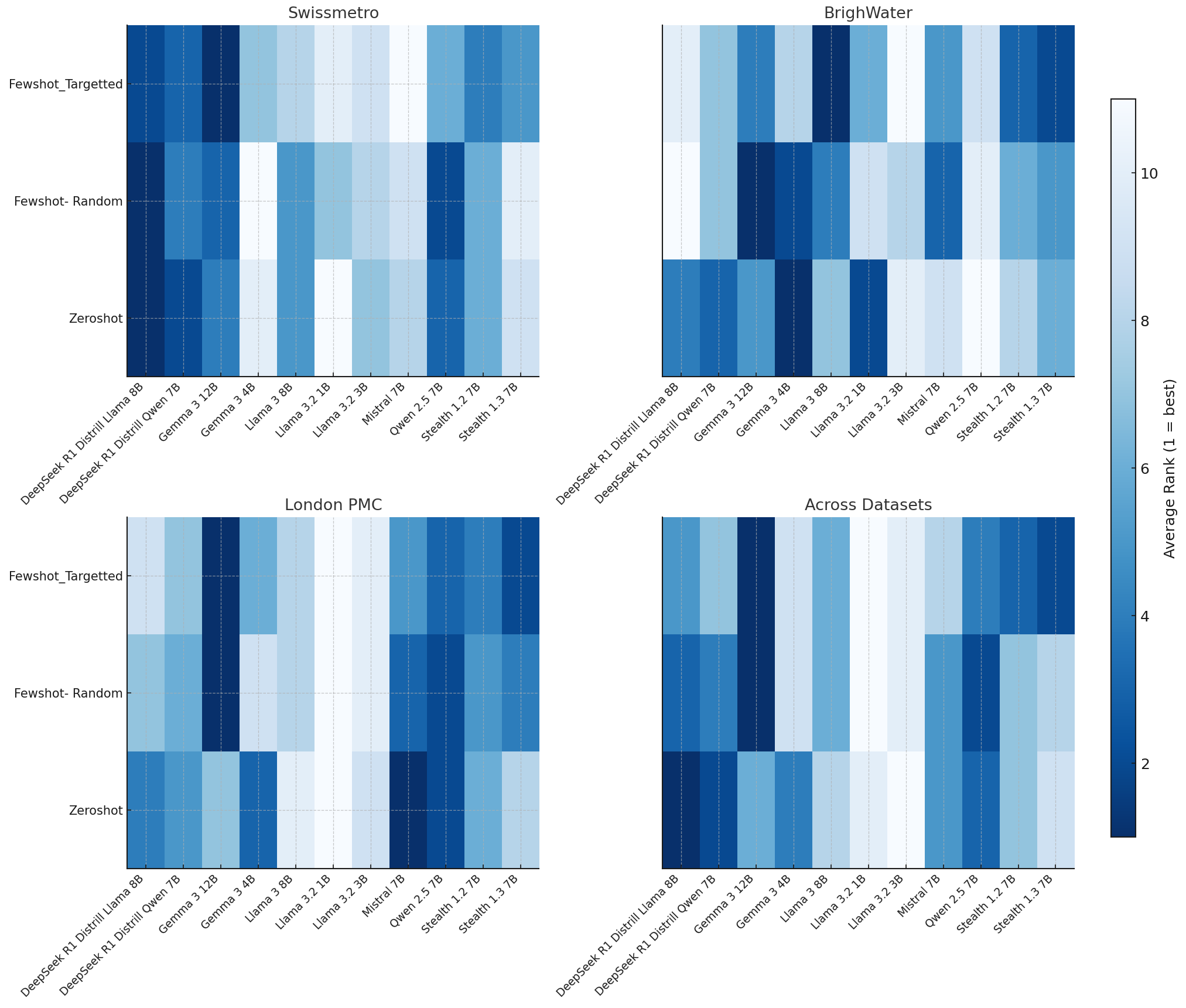}
    \caption{Model's Predictive Performance Ranking}
    \label{fig: Models Predictive Performance Ranking}
\end{figure}

With the model selection emerging as the dominant source of variation in predictive performance across datasets (Figure~\ref{fig: Variance Explanation}), it warrants a closer examination of individual model behaviour and their relative performance. Specifically, we rank the models based on their achieved weighted F1 scores under the three learning styles across the different travel survey datasets. Figure~\ref{fig: Models Predictive Performance Ranking} presents the aggregated ranks for each model, averaged across prompt styles and temperature settings for each configuration.
For zero-shot settings, where no contextual examples are provided to guide inference, DeepSeek R1 Distill Llama 8B and DeepSeek R1 Distill Qwen 7B consistently occupy the top ranks. Both models are distilled versions of the larger DeepSeek R1 reasoning model, employing reinforcement learning during distillation to preserve chain-of-thought capabilities in compact architectures of seven to eight billion parameters. Their performance proves how well designed distillation processes can embed advanced reasoning into lightweight, deployable models, making them highly suitable for tasks with no annotated demonstrations, and where minimum contextual guiding examples are available.

When random few-shot prompting is introduced, Gemma 3 12B emerges as the top performer. This large scale transformer leverages a hybrid attention mechanism that alternates between local and global attention blocks, allowing it to process long sequences up to 128,000 tokens without loss of coherence. Even with uncurated examples, the model shows superior contextual absorption, significantly outperforming others. Nonetheless, the DeepSeek distillates and Qwen 2.5 7B remain competitive, mainly due to the advantage of combining strong reasoning priors with scalable context length.

In the targeted few-shot, where training examples are carefully selected based on similarity to the test case, Gemma 3 12B further consolidates its lead. Its architecture proves particularly effective at internalizing structured demonstrations when available. Notably, Stealth 1.2 7B and Stealth 1.3 7B also rise in rank under this setting. These models, although less documented, are optimized for local inference and appear to have benefited from finetuning on diverse instructional datasets, allowing them to respond well to high quality contextual cues.

This ranking analysis helps showcase the importance of selecting models not solely by parameter count, but by architectural features and training history. The distilled DeepSeek variants offer high accuracy at low computational cost in zero-shot settings. Mid-sized dense models such as Qwen 2.5 7B and Mistral 7B provide a reliable trade-off between generalizability and efficiency, while larger models like Gemma 3 12B yield significant returns when prompted with curated examples. Overall, the results indicate that the optimal model choice depends critically on the deployed learning style and the availability of demonstration examples.

\subsubsection{ Learning Style and Learning Potential of LLMs for Predictive Tasks}
\begin{figure}[!ht] 
    \centering
    \includegraphics[width={\linewidth}]{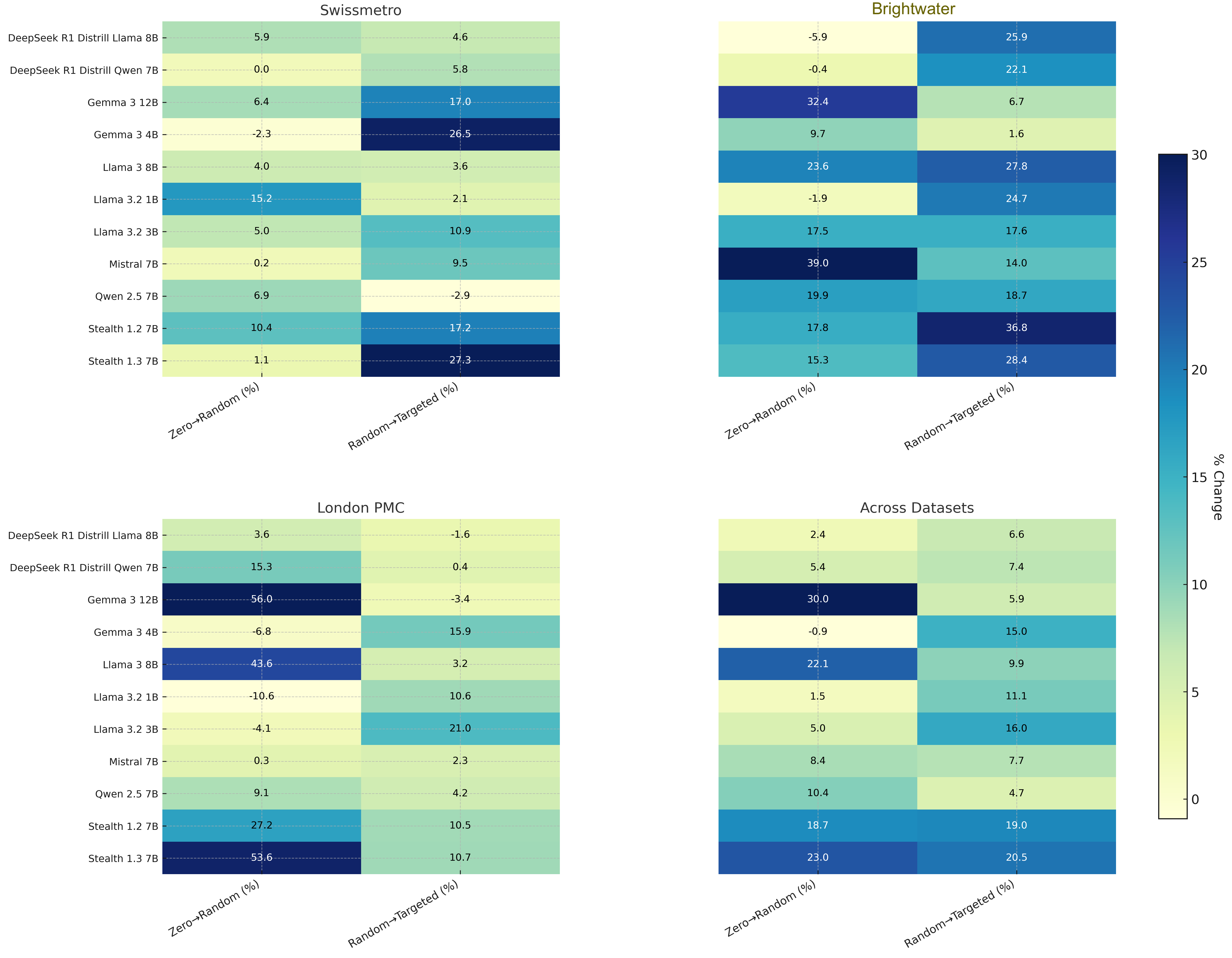}
    \caption{Impact of Learning Style on the Model's Predictive Performance Gain(weighted F1 Change)}
    \label{Impact of Learning Style}
\end{figure}

Figure~\ref{Impact of Learning Style} shows for each model and dataset the percentage change in weighted F1 when moving from zero-shot to few-shot random and from few-shot random to few-shot targeted. These relative gains expose each model’s capacity to learn from uncurated versus carefully chosen in-context examples.

In the Swissmetro survey, most models improve by 4--15\% when given random examples and by an additional 2--27\% under targeted examples. Notably, Gemma 3 4B loses performance (--2.3\%) with random prompts, a clear case of negative learning likely caused by unrepresentative support sets demonstrated by the strong rebound when supplied with curated examples. By contrast, the Stealth 1.2 7B and Stealth 1.3 7B models display substantial random-shot gains and the largest targeted-shot uplifts (17.2\%, 27.3\%), highlighting their surprising adaptability and learning potential despite modest parameter counts and baseline performance.
On the BrightWater dataset, which is characterized by high behavioural heterogeneity, Mistral 7B leads all models with a random-shot gain of 39.0\%, while Stealth 1.2 7B peaks under targeted examples at 36.8\%. Only the distilled variants of DeepSeek R1 show slight declines (--5.9\%, --0.4\%) with random sampling as a result of uncurated examples, which might misalign with their fixed reasoning priors. However, once examples are curated, even these distillates recover and show substantial gains of 22.1--25.9\%.
For the London PMC dataset, where zero-shot baselines are already strong, Gemma 3 12B exhibits the most dramatic learning curve, with a 56.0\% improvement under random examples, but also the largest volatility, dropping -3.4\% when switching to targeted. Smaller LLaMA variants, such as Llama 3.2 1B, show negative learning under random (--10.6\%) but rebound (+10.6\%) with targeted examples, these models lightweight architectures appear more sensitive to example quality but can still benefit from well-aligned support sets.

When averaged across all surveys, the stealth models stand out with the largest sequential gains (18.7, 23\%) from zero to random, and an additional 19\% to 20.5\% from random to targeted, combining strong learning potential with reduced negative learning. On the other hand, the distilled models post modest average improvements (2.4--5.4\% from zero to random, 6.6--7.4\% from random to targeted), mainly as a result of their limited plasticity once reasoning patterns are fixed during distillation.

These findings support three key insights. First, negative learning can occur when support examples conflict with pretrained expectations or when selected samples are largely misaligned with the test case, particularly in smaller models. Second, instruction-tuned, mid-sized models like those from the Stealth family offer a strong adaptability and learning potential with substantial gains once presented with in-context examples. Third, large, dense transformers such as Gemma 3 12B can leverage random support sets effectively, and learn the underlying relationship between the commuters' profile and their mode choice, which can be further improved with curated examples.
In practice, deploying a few random-shot learning style without quality control should be avoided, especially in complex or noisy environments. Where curated examples are available, their impact can be substantial. Among the tested models, Stealth and Gemma variants emerge as robust and versatile learners, while distilled models remain a reliable choice for zero-shot deployment when examples cannot be supplied.

\subsubsection{Prompting Style and Model's Temperature Performance Impact}
\begin{figure}[!ht]
    \centering
    \includegraphics[width=\linewidth]{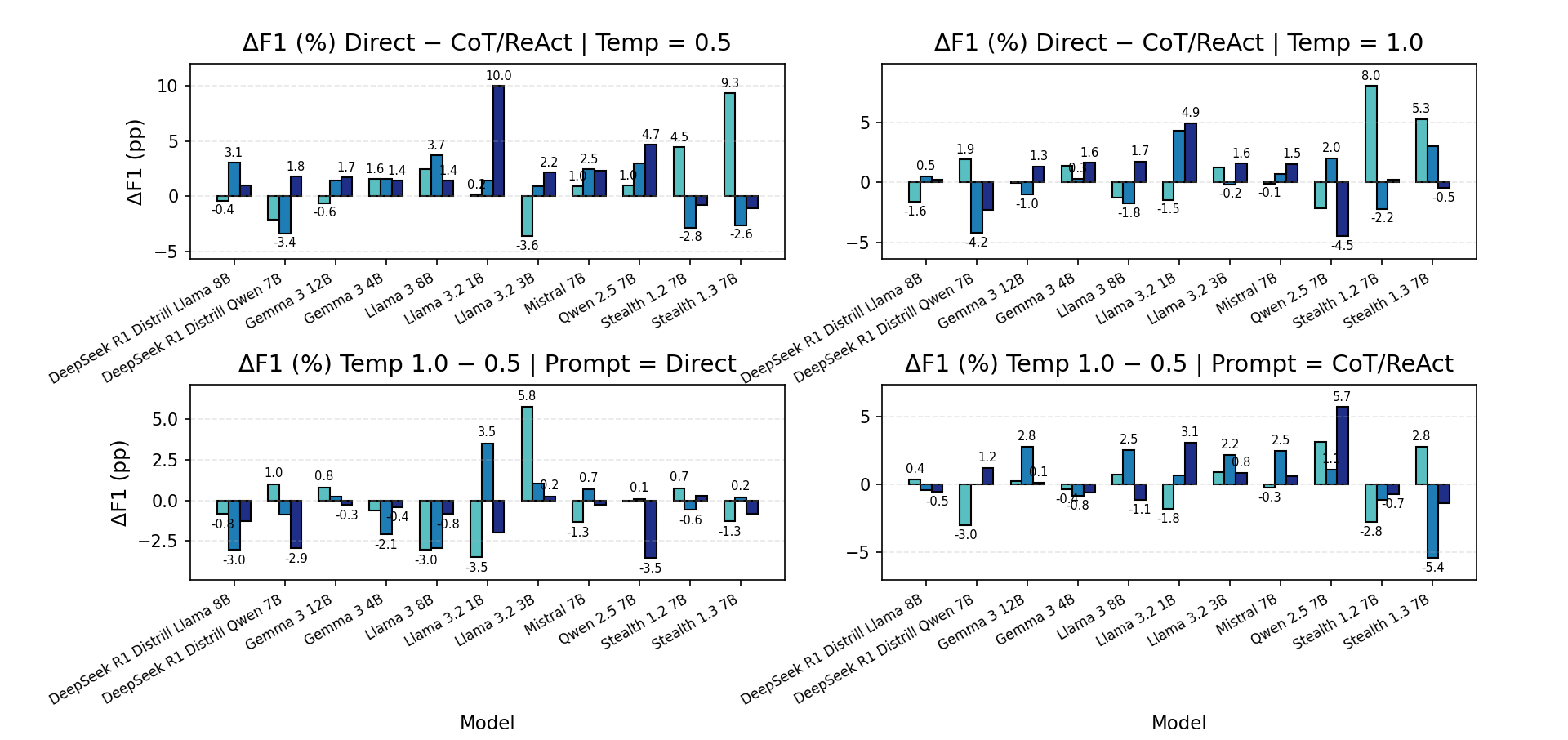}
    \caption{Impact of Prompt Style and Model Temperature on the Model Predictive Power}
    \label{fig:prompt-temp-impact}
\end{figure}

Building on the earlier findings regarding model architecture, dataset characteristics, and shot styles, we now turn to a finer grained examination of two additional dimensions influencing predictive performance: the prompt style (Direct vs.\ Chain-of-Thought/ReAct) and the model temperature (0.5 vs.\ 1). These two axes offer insight into how LLMs respond to adjustments in prompt structure and sampling stochasticity, independent of parameter count or dataset complexity. These elements, while often overlooked, introduce subtle but meaningful shifts in model behaviour across the three travel survey datasets.
Figure~\ref{fig:prompt-temp-impact} shows, for each model across the survey dataset, the percentage point change in weighted F1 score when switching from Chain-of-Thought/ReAct to Direct prompting and from temperature 0.5 to temperature 1, across all three shot styles: zero-shot, few-shot random, and few-shot targeted.

To disentangle the independent effects of prompting style and decoding temperature, we computed four contrasts while holding the other factor constant: (i) $\Delta F_{1}^{\text{prompt}}=\text{F1}_{\text{Direct}}-\text{F1}_{\text{CoT/ReAct}}$ at $T=0.5$, (ii) the same prompt contrast at $T=1.0$, (iii) $\Delta F_{1}^{\text{temp}}=\text{F1}_{T=1.0}-\text{F1}_{T=0.5}$ with the prompt fixed to \textit{Direct}, and (iv) the same temperature contrast with the prompt fixed to \textit{CoT/ReAct}. 

Across the board, differences arising from prompt style were larger and more systematic than those from temperature changes. Switching from a Chain-of-Thought/ReAct prompt to a Direct prompt generally improved performance, especially once illustrative examples were available. This effect was most pronounced in the Fewshot\_Targetted learning style, where Direct prompting consistently sharpened the model's focus. For instance, with temperature held at 0.5, using a Direct prompt in the Fewshot\_Targetted setting resulted in positive mean F1 score changes across all three datasets (Brightwater: +1.58\%, London PMC: +4.63\%, Swissmetro: +0.52\%) and a positive impact on the weighted F1 score in 9 out of 11 models. However, results in the Zeroshot were noisier and more model dependent; while most models (7) still benefited from concise Direct prompts, others (4) performed better with the reasoning structure provided by CoT/ReAct. These patterns confirm that prompt format is not a one-size-fits-all hyperparameter but must be aligned with the model’s architecture and the specific task context.

Temperature adjustments (from 0.5 to 1.0) produced more subtle effects. Contrary to improving performance, increasing the temperature to 1.0 often led to negligible changes or slight performance declines, especially under a direct prompting style. This finding indicates a general preference for the more deterministic decoding associated with a temperature of 0.5. For example, when using Direct prompting, the number of models that lost performance at $T=1.0$ was greater than the number that gained performance across all datasets and shot types. A critical finding is the non-trivial interaction between temperature and prompt style. The effect of changing temperature often diverged in both magnitude and direction depending on the prompt. For the Qwen 2.5 7B model, increasing temperature under a Direct prompt degraded performance by $-3.5$\% for few-shot targeted, whereas under a CoT/ReAct prompt, it improved performance by + 5.7\%.

Architectural and training differences help explain these heterogeneous responses. Dense instruction-tuned models (e.g., the Gemma, Llama, and Mistral families) typically capitalised on Direct prompts, especially in the Fewshot\_Targetted setting, which is more of a reflection of their fine-tuning on concise task specifications. Larger models, like Gemma 3 12B model, although showed consistent improvement with Direct prompting at $T=0.5$, it was overall less sensitive to varying prompting styles or temperatures. On the other hand, smaller models showed extreme sensitivity, with the Llama 3.2 1B model showing a massive +30.37\% F1 score increase on the London PMC dataset under these conditions. On the other hand, certain distilled reasoning models, such as DeepSeek R1 Distill Qwen 7B, proved more fragile to prompt compression, often losing accuracy when the CoT/ReAct reasoning chain was removed. These models were also more sensitive to increases in temperature, which often introduced harmful variability.

In practical terms, these results suggest a clear two step tuning strategy for the mode choice predictive task. First, one should calibrate the prompt style (Direct vs. CoT/ReAct) under a fixed, conservative temperature (e.g., 0.5), as this choice exerts the dominant influence on performance. Second, only after the optimal prompt style is determined for the target learning style should one probe higher temperatures to see if mild stochasticity (e.g., $T=1.0$) offers any reproducible gains. This held-constant analysis clarifies that hyperparameters cannot be tuned in isolation; their effects are coupled, and their optimal settings depend on the underlying model, the number of examples provided, and the dataset itself.

\subsubsection{Comparative Performance on Stated vs. Revealed Preference Datasets} 

To examine whether LLMs exhibit different learning behaviour when exposed to revealed versus stated preference data, we compare their performance across the Swissmetro and BrightWater stated-preference (SP) datasets and the London PMC revealed-preference (RP) dataset. Across all learning styles, models consistently attain higher weighted F1 and accuracy on the RP dataset, despite its smaller training pool for few-shot examples. Under targeted few-shot prompting, London PMC achieves a mean weighted F1 exceeding 0.67, compared with 0.61 for Swissmetro and 0.58 for BrightWater. Even in the zero-shot setting, the RP dataset produces stronger baselines, indicating that LLMs can infer realistic mode-choice behaviour with minimal contextual guidance. This contrast suggests that causal LLMs generalize better from real-world behavioural signals than from hypothetical survey data. In RP data, observed time–cost trade-offs, accessibility variables, and modal separability provide clear decision boundaries that the model can internalize even from a limited number of examples to choose from. In SP data, however, the presence of hypothetical or aspirational choices introduces additional noise and weakens the relative importance of explanatory features, leading to greater variability and lower per-instance discrimination.

The higher predictive stability of the RP dataset, achieved despite a smaller training pool for few-shot examples, indicates that LLMs can capture genuine travel behaviour more effectively than self-reported intentions. This offers a new perspective on mitigating hypothetical bias in travel survey based modelling, which often limits the realism and practical utility of SP data for planning purposes.

\subsection{Distribution-Level LLMs Mode Choice Predictive Performance Evaluation}
\begin{figure}[!ht]
    \centering
    \includegraphics[width=\linewidth]{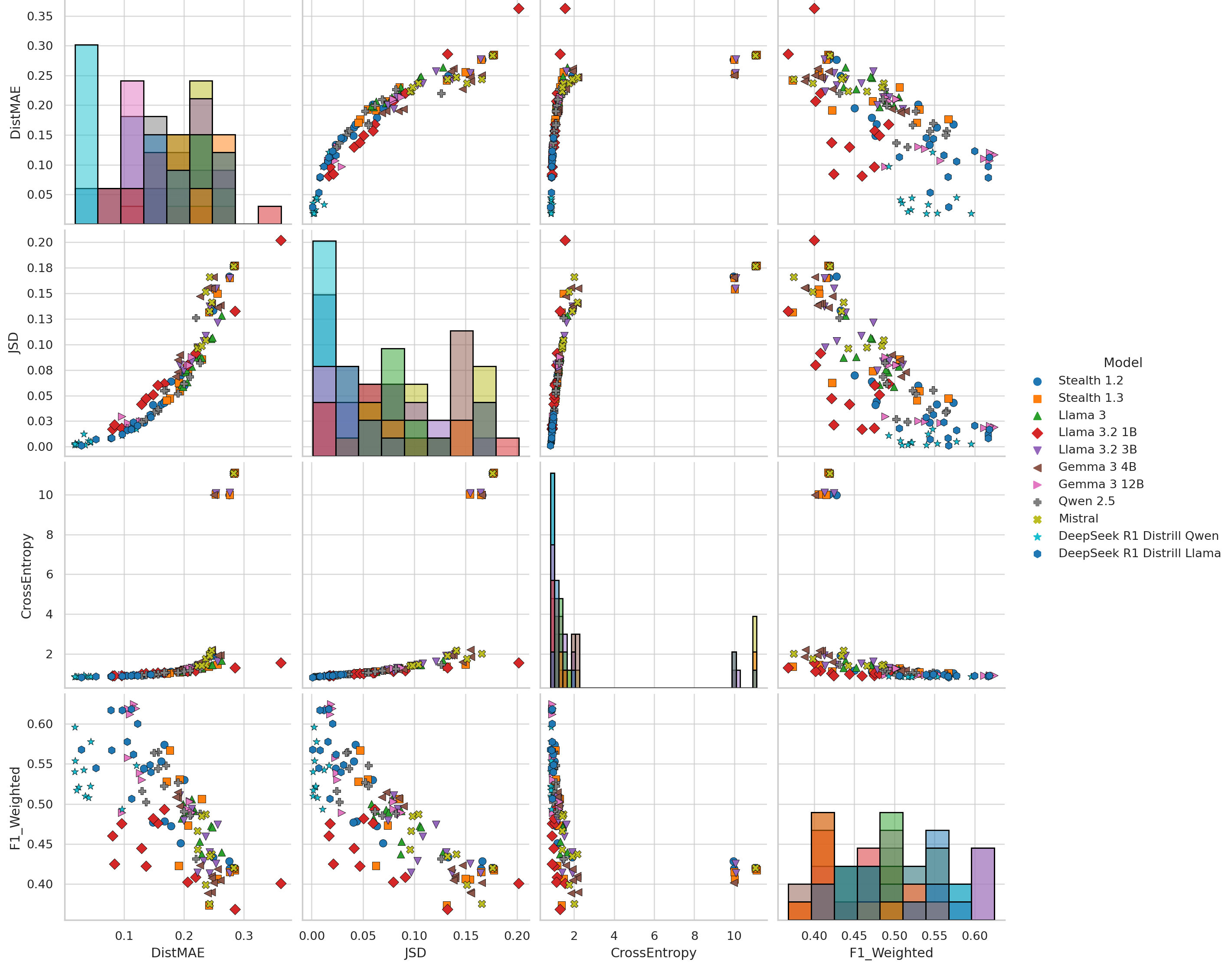}
    \caption{Pairwise Scatter of LLMs predictive Performance at Mode Choice Distribution Vs Instant Level Prediction- Swissmetro Dataset}
    \label{fif:paiwise_swissmetro}
\end{figure}

\begin{figure}[!ht]
    \centering
    \includegraphics[width=\linewidth]{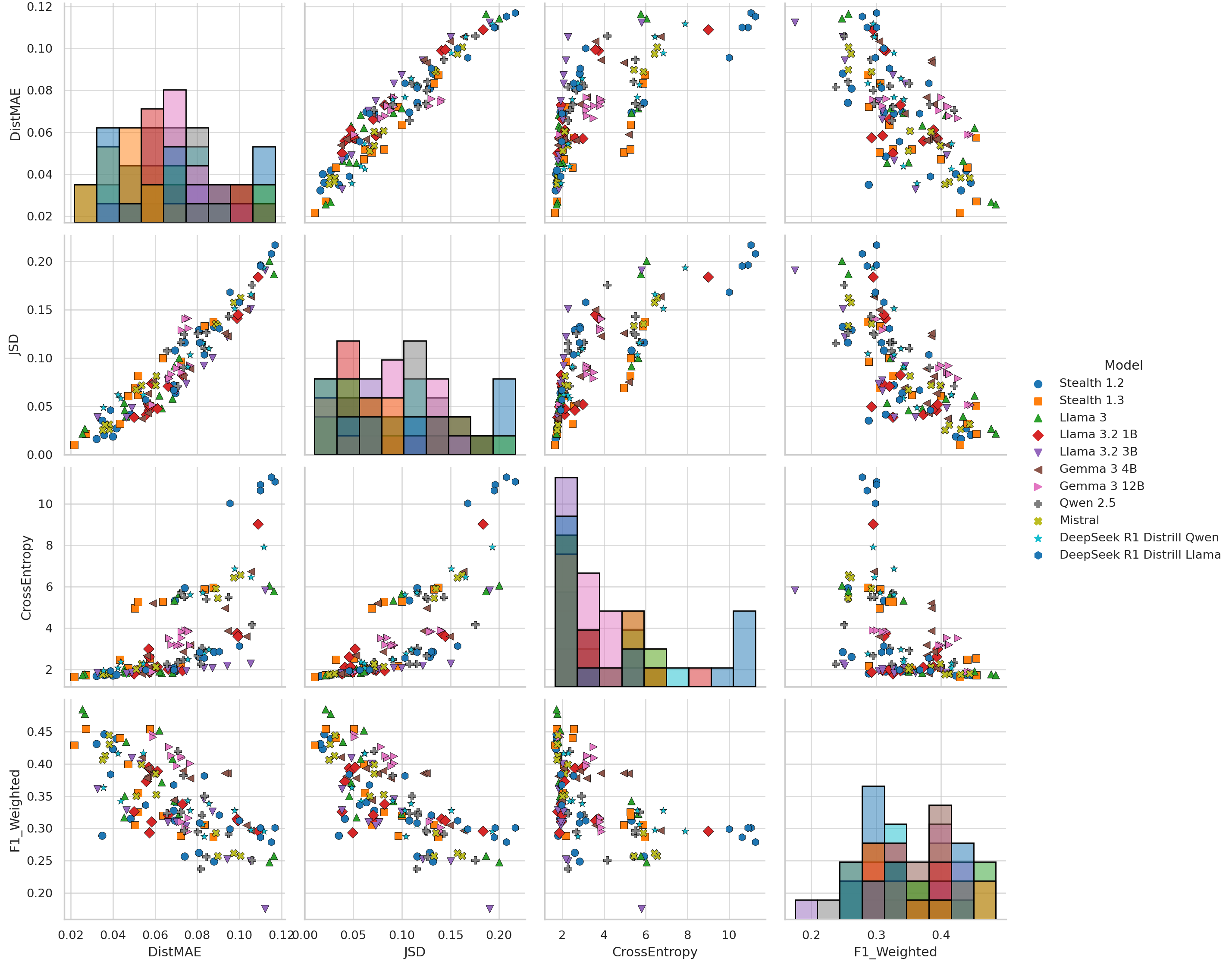}
    \caption{Pairwise Scatter of LLMs predictive Performance at Mode Choice Distribution Vs Instant Level Prediction- Brightwater Dataset}
    \label{fif:paiwise_brightwater}
\end{figure}

\begin{figure}[!ht]
    \centering
    \includegraphics[width=\linewidth]{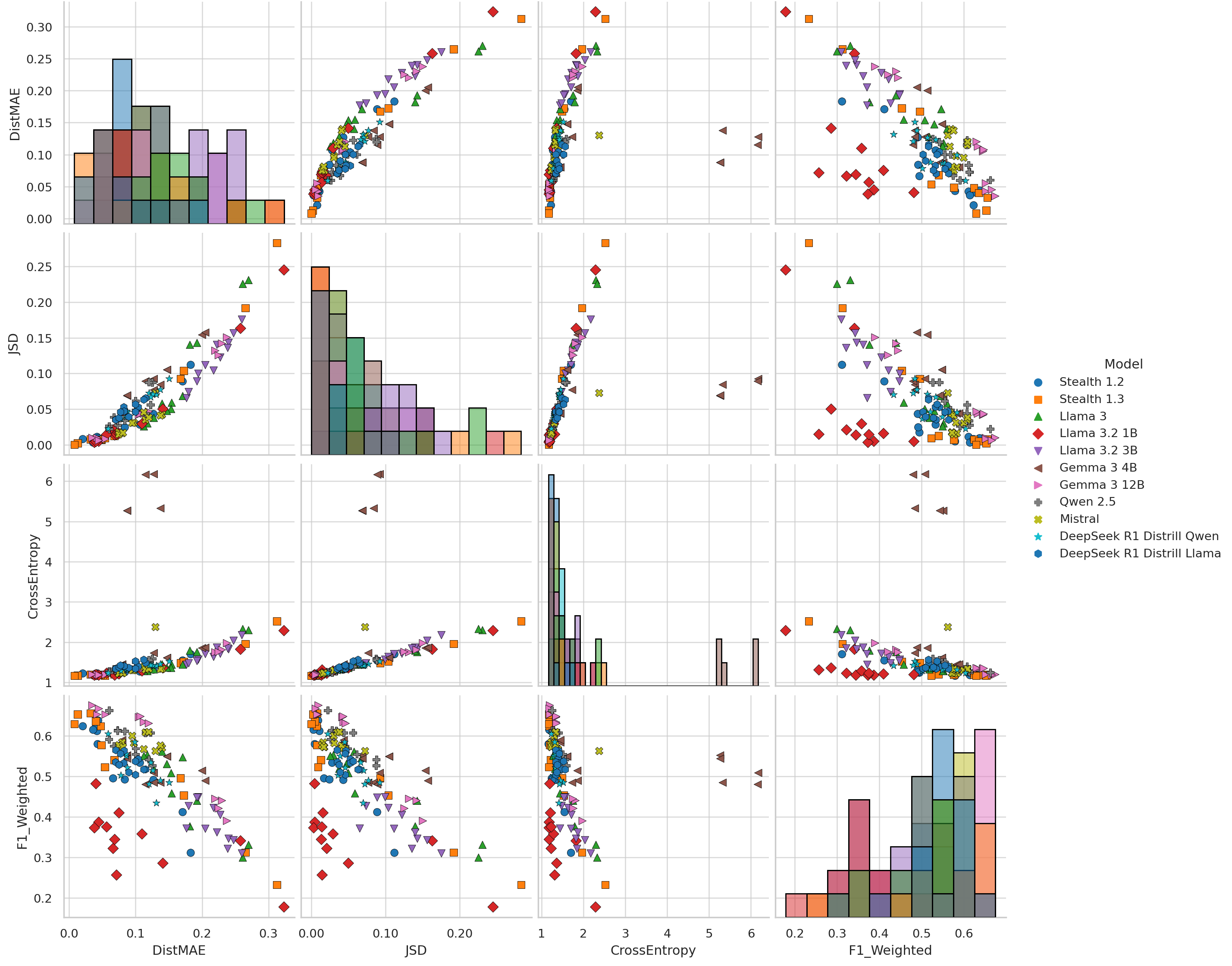}
    \caption{Pairwise Scatter of LLMs predictive Performance at Mode Choice Distribution Vs Instant Level Prediction- London PMC Dataset}
    \label{fif:paiwise_london}
\end{figure}

In addition to evaluating predictive performance at the instance level, we assess how well each model captures the overall structure of travel behaviour by examining performance at the distribution level. Specifically, we compare the predicted choice distributions against the empirical distributions observed in the survey data, quantifying divergence using three complementary metrics: Jensen–Shannon Divergence (JSD), Mean Absolute Error (MAE), and Cross Entropy. Figures~\ref{fif:paiwise_swissmetro}, \ref{fif:paiwise_brightwater}, and \ref{fif:paiwise_london} present pairwise scatter plots showing how each model performs on these distributional metrics relative to its instance-level weighted F1 score, across the three datasets.

Across all datasets, DistMAE and JSD track each other closely because both measure how well the aggregate selection shares align with the empirical mode shares. Cross-entropy also correlates with them, but it penalises omissions much more strongly: when a present class receives zero or near-zero aggregate share in a run, the loss spikes, producing pronounced high outliers.

For the Swissmetro dataset (Figure \ref{fif:paiwise_swissmetro}), models such as DeepSeek R1 Distill Llama 8B, Gemma 3 12B, and DeepSeek R1 Distill Qwen 7B achieve high instant-level accuracy and low distribution-level error. Their selections keep minority modes represented, so no class collapses to a zero count and cross-entropy remains low. Weaker models such as Mistral 7B and Stealth 1.3 7B show volatile cross-entropy with visible outliers. In those runs, the selections concentrate too heavily on the dominant mode and leave one or more valid classes with zero selections.

Brightwater shows a distinct pattern. Models generally achieve lower instant-level accuracy, yet the distribution-level picture is mixed: DistMAE sits in a tight band (about 0.02–0.12), while JSD and cross-entropy vary widely, with cross-entropy spanning roughly 1.6–11 (Figure \ref{fif:paiwise_brightwater}). Two structural features explain this. First, the choice set contains nine dynamic alternatives, several with small shares. With many thin tails, absolute deviations accumulate slowly: a model can mis-allocate small amounts across rare modes and still keep the sum of absolute errors modest, hence the low DistMAE. JSD responds to relative error: a shift from 0.01 to 0.03 is small in absolute terms but large in relative terms, so JSD rises even when DistMAE stays low. Cross-entropy is stricter still: if a run assigns zero selections to a present class, the loss spikes. Second, the respondent population is comparatively homogeneous. Although the model never observes the aggregate distribution, many instances share similar covariates, so models tend to produce similar selections across cases. When aggregated, the resulting shares sit close to the empirical shares (low DistMAE) but per-instance discrimination remains modest (lower F1). 

London PMC achieves the highest instant-level accuracy, with moderate distribution errors and few cross-entropy outliers (Figure \ref{fif:paiwise_london}). Three features contribute. First, the option set contains fewer very low-share classes than Brightwater, so tail-risk omissions are rarer and cross-entropy is steadier. Second, covariates separate the dominant mode clearly, time–cost trade-offs and access variables support a firmer decision boundary, which lifts per-instance accuracy and reduces the need to spread selections thinly. Third, the aggregate shares are stable enough that few-shot exemplars provide useful priors; random few-shot already attains high F1, while targeted few-shot further improves calibration by lifting minority-mode counts and lowering JSD with only small changes in F1. Despite these advantages, the distribution metrics do not collapse to a floor: DistMAE and JSD remain mid-range because high-accuracy models still concentrate selections on the predicted winner and trim secondary modes. Qwen 2.5 7B and Mistral 7B balance this well, sitting near the top in F1 while keeping DistMAE and JSD low. Stealth 1.2 7B offers a complementary profile: slightly lower F1 but the best DistMAE and among the lowest cross-entropy, indicating aggregate shares close to the ground truth. Gemma 3 12B reaches very high F1 under fewshot–random but often places too much weight on the chosen alternative, which in some cases lifts DistMAE and JSDs. 

Discrepancies between instant- and distribution-level metrics are expected. A model can score highly on weighted F1 by excelling on majority classes while under‑selecting minority modes. This elevates JSD and, when a present class receives zero selections, produces large cross‑entropy. Conversely, a model that disperses selections more evenly may forgo some top‑1 hits (lower weighted F1) yet produce aggregate shares closer to the truth, which lowers DistMAE. The prompting strategy controls this balance at the instance level. Targeted few‑shot steers selections toward locally plausible alternatives; F1 improves and minority‑mode counts rise, which usually lowers JSD and cross‑entropy. Random few‑shot improves F1 over zero‑shot but can still misalign selections with the test context, leaving JSD elevated and occasionally producing cross‑entropy spikes. Zero‑shot yields both lower F1 and weaker calibration.

\begin{figure}[!ht]
  \centering
  \includegraphics[width=\linewidth]{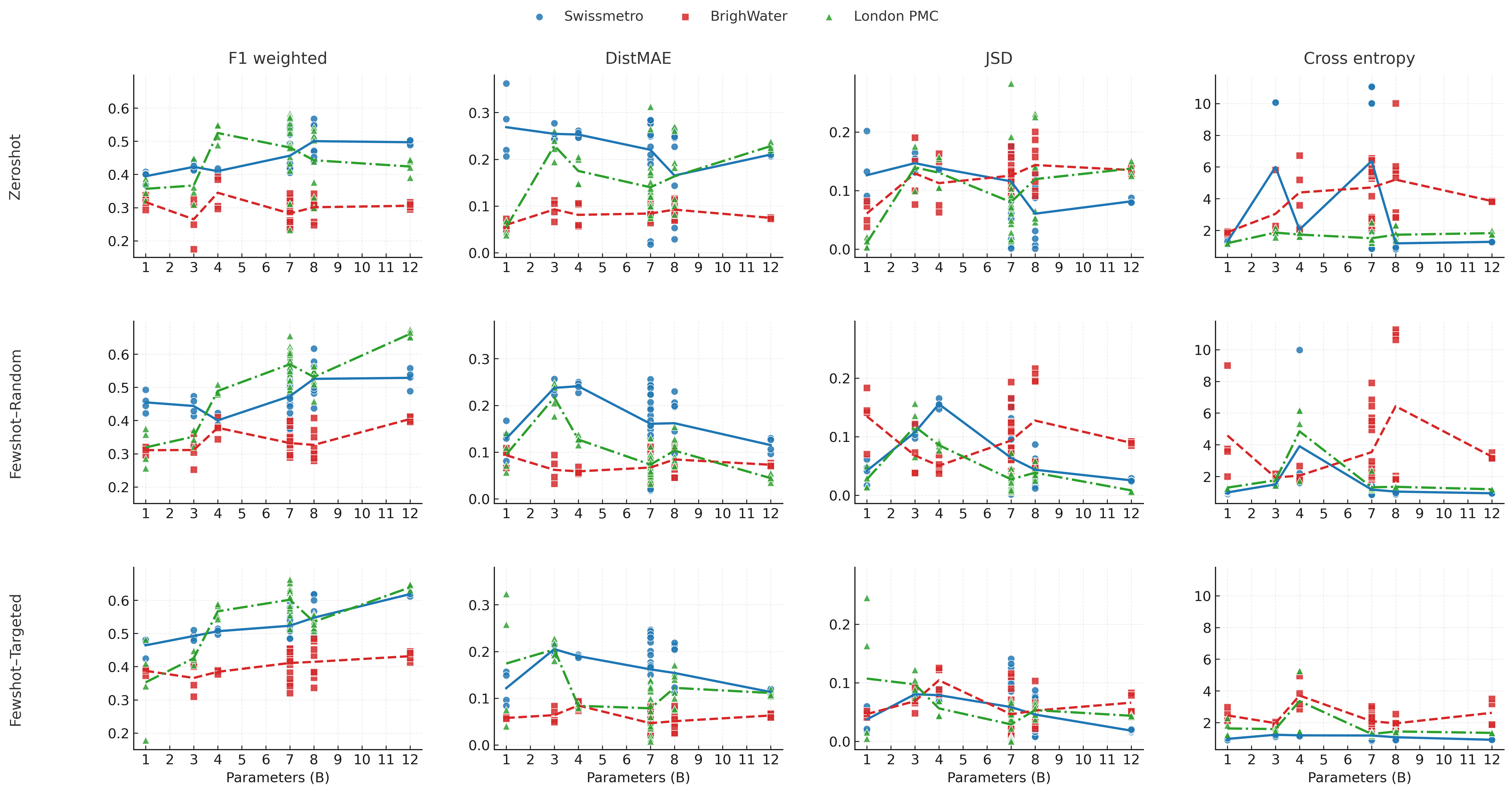}
  \caption{Model size (parameters in billions) versus instant- and distribution-level metrics across all datasets and shot types}
  \label{fig:size_vs_metrics_all}
\end{figure}

In terms of the Model size effects, Figure~\ref{fig:size_vs_metrics_all} and Table~\ref{tab:size_trends} jointly summarize how parameter count relates to performance. The figure shows points for individual runs and per‑dataset mean trends at each parameter size. The table encodes the dominant direction of change with arrows: up ($\uparrow$), down ($\downarrow$), approximately flat ($\rightarrow$); slanted arrows ($\nearrow,\,\searrow$) indicate weaker tendencies.

\begin{table}[ht]
\centering
\footnotesize
\setlength{\tabcolsep}{6pt}
\begin{tabular}{lcccc}
\toprule
Dataset \quad \quad \quad Shot & F1 weighted & DistMAE & JSD & Cross entropy \\
\midrule
Swissmetro \quad  Zeroshot          & $\uparrow$    & $\downarrow$ & $\downarrow$ & $\searrow$ \\
Swissmetro \quad Fewshot--Random   & $\uparrow$    & $\rightarrow$/ $\searrow$ & $\searrow$   & $\rightarrow$ \\
Swissmetro \quad Fewshot--Targeted & $\nearrow$    & $\rightarrow$/ $\searrow$ & $\rightarrow$/ $\searrow$ & $\rightarrow$ \\
\addlinespace
BrigthWater \quad Zeroshot          & $\rightarrow$ & $\rightarrow$ & $\uparrow$ & $\uparrow$ \\
BrigthWater \quad Fewshot--Random   & $\rightarrow/\nearrow$ & $\rightarrow$ & $\nearrow$ & $\rightarrow/\nearrow$ \\
BrightWater \quad Fewshot--Targeted & $\rightarrow/\nearrow$  & $\rightarrow$ & $\rightarrow$/ $\searrow$ & $\rightarrow$ \\
\addlinespace
London PMC \quad Zeroshot          & $\nearrow$    & $\nearrow$  & $\nearrow$ & $\rightarrow$ \\
London PMC \quad Fewshot--Random   & $\uparrow$   & $\downarrow$ & $\downarrow$ & $\searrow$ \\
London PMC \quad Fewshot--Targeted & $\uparrow$  & $\searrow$ & $\rightarrow$/ $\searrow$ & $\rightarrow$/ $\searrow$ \\
\bottomrule
\end{tabular}
\caption{Direction of change with increasing parameter count, by dataset and shot type. Up $(\uparrow)$, down $(\downarrow)$, flat $(\rightarrow)$; slanted arrows $(\nearrow,\,\searrow)$ denote weaker trends.}
\label{tab:size_trends}
\end{table}

Three mechanisms explain the patterns. First, separation versus coverage. On Swissmetro, stronger covariate signals reward decisive separation; larger models raise F1 and still keep minority modes represented, so DistMAE and JSD decline or stabilise at low levels. Once few‑shot context is provided, divergence varies less across sizes and further gains are small. On Brightwater, with nine alternatives and several rare modes, size without context often hardens decisions on dominant classes. That can nudge F1 up while under‑selecting rare modes, lifting JSD and, when a present class is omitted, cross entropy. Targeted prompts counter this by anchoring selections to locally plausible options; with that anchor, larger models improve both F1 and calibration. London PMC lies between the two: the task is easier than Brightwater, so even mid‑size models achieve good accuracy. Larger models still help, but the marginal reduction in divergence is gentle and cross entropy is already low because tail‑risk omissions are rare.

Second, interaction with learning style. Zeroshot exposes raw inductive bias: when the class structure is skewed (Brightwater), larger models amplify that bias unless guided. Random few‑shot often lifts F1 but does not guarantee coverage of rare modes, so JSD and cross entropy can remain elevated. Targeted few‑shot is the setting where scale reliably converts to better calibration: minority‑mode counts rise, DistMAE and JSD tilt downward, and cross‑entropy spikes disappear.

Third, training recipe versus size. Differences among families matter as much as parameter count on London PMC and, to a lesser extent, on Swissmetro under few‑shot. Qwen 2.5–7B and Mistral 7B achieve high F1 with steady calibration across shots. Reasoning‑distilled models such as DeepSeek R1 Distill models (Llama 8B, Qwen 7B) often have an edge at the same parameter scale on Swissmetro: they separate alternatives cleanly while maintaining coverage of minority modes, yielding high F1 with low DistMAE and JSD and few cross‑entropy outliers. On Brightwater the same decisiveness can overshoot in zeroshot or random few‑shot; targeted prompting restores balance and the models then sit among the best calibrated for their size. On London PMC they remain competitive but not uniformly superior: Qwen 2.5 7B and Mistral 7B deliver comparable accuracy with stable divergence, suggesting instruction tuning shapes performance at least as strongly as parameter count.

Overall, model size improves instant‑level prediction when the dataset offers clear separating structure or when prompts supply that structure. For distribution‑level performance, larger models help once minority‑mode coverage is ensured, either inherently (Swissmetro) or via targeted prompting (Brightwater). Where the task is already well separated (London PMC), training recipe and shot type are the primary levers; size provides incremental gains.

From a practical perspective, accurate prediction of mode shares is crucial for forecasting applications, requiring models with robust calibration at the distribution level and high instant-level accuracy. Models responsive to targeted few-shot prompting are good candidates for supervised fine‑tuning, provided prompting examples are carefully selected to match the dataset and model. Overall, the pairwise analysis shows the balance between instant-level accuracy and distribution-level calibration to aid the selection models that align with the intended application.

\subsection{Reasoning Analysis: Topics, Factor Coverage, and Semantic Structure}

For this analysis, we focus exclusively on the Swissmetro dataset. This decision is motivated by its extensive use in behavioural modelling and its wide adoption as a benchmark in recent studies exploring LLMs for travel mode choice inference \citep{liu2025aligning, mo2023large}. Swissmetro provides a controlled setting with a clear set of alternatives and well-documented trade-offs, making it particularly suitable for examining how LLM-generated reasoning aligns with choice semantics. 

\begin{figure}[ht!]
\centering
\includegraphics[width=\linewidth]{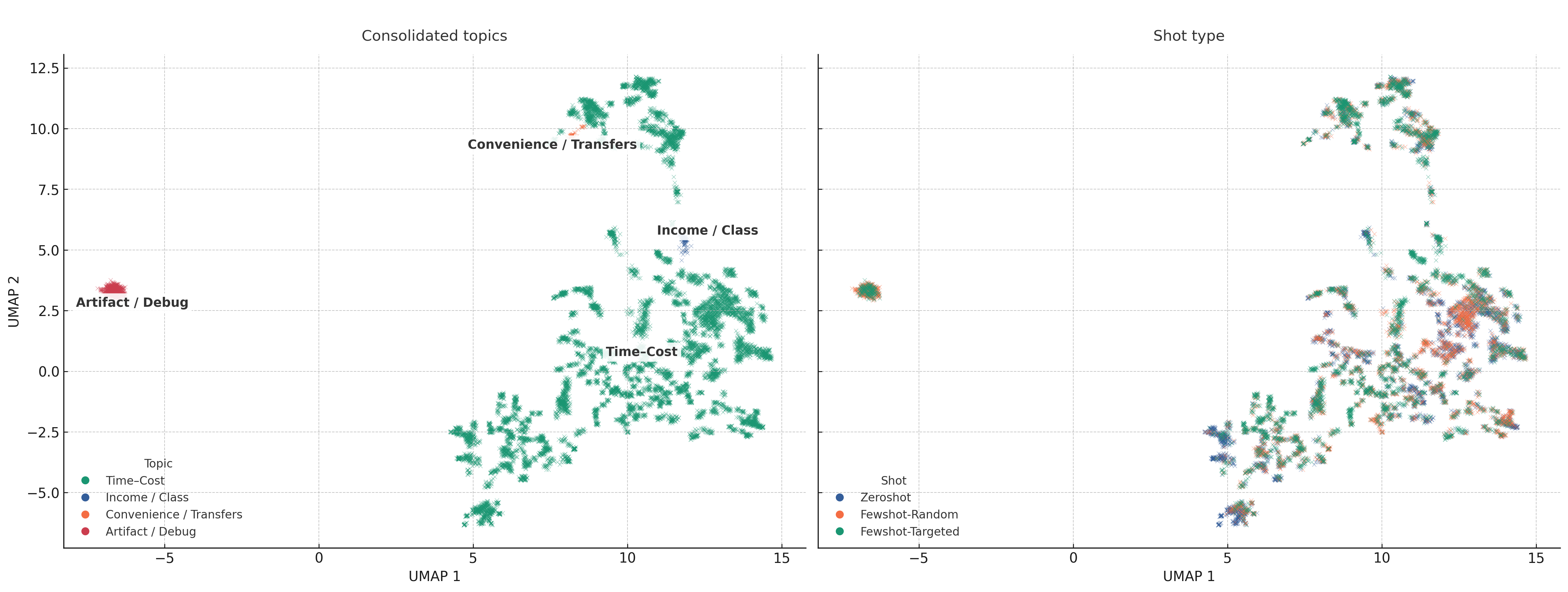}
\caption{UMAP projection of Swissmetro reasoning texts. Left: consolidated topic labels; right: shot type}
\label{fig:sm_umap_dual}
\end{figure}

The Swissmetro reasoning corpus resolves into a dominant family that weighs travel time against cost. After strict cleaning and BERTopic modelling, several numerical topics carry the same semantics; we therefore consolidate them under a single label (Time--Cost) and retain three smaller, semantically distinct themes: Income / Class, Convenience / Transfers, and a residual Artifact / Debug that captures prompt remnants and formatting noise. The consolidated geometry and the shot–type overlay appear in Figure~\ref{fig:sm_umap_dual}. Points that lie close in the UMAP share vocabulary and phrase structure, while separation indicates distinct justificatory semantics.

Figure~\ref{fig:sm_umap_dual}, left panel shows how a large, contiguous region corresponds to Time--Cost. Its footprint is not uniform: a dense central spine contains short, decisive sentences contrasting total times and costs; an upper arc contains longer phrases that balance time savings against cost differences and occasionally mention comfort or reliability; a lower arc aligns with formulations framed against the train alternative. These internal pockets explain why multiple BERTopic indices mapped to the same label: they represent sub-themes within one argumentative family rather than genuinely different topics. Two compact satellites sit adjacent to the main mass. Income / Class gathers references to income, first-class seating, and season-ticket affordability; lexical specificity pulls these statements together and away from the general trade-off cluster. Convenience / Transfers groups mentions of transfers, connections, and door-to-door practicality; the cluster remains small on Swissmetro and lies near the time–cost boundary because many texts still include an explicit time reference. The Artifact / Debug cluster sits well apart on the left, confirming that residual boilerplate has been successfully quarantined and does not contaminate the substantive clusters. 

Shot type modulates occupancy more than geometry (Figure~\ref{fig:sm_umap_dual}, right panel). Targeted few-shot concentrates points along the central Time--Cost spine and increases occupancy in the two satellites, most visibly in Income / Class. Random few-shot spreads points more broadly across the main region. Zeroshot populates the periphery and rarely enters the satellites. This pattern aligns with the distribution metrics: targeted prompts steer statements toward locally decisive cues and add just enough secondary evidence (ticket ownership, transfer burden) to preserve minority alternatives, which reduces JSD and prevents cross-entropy spikes while maintaining, and in several cases improving, weighted F1 on Swissmetro.

\begin{figure}[!t]
\includegraphics[width=0.9\linewidth]{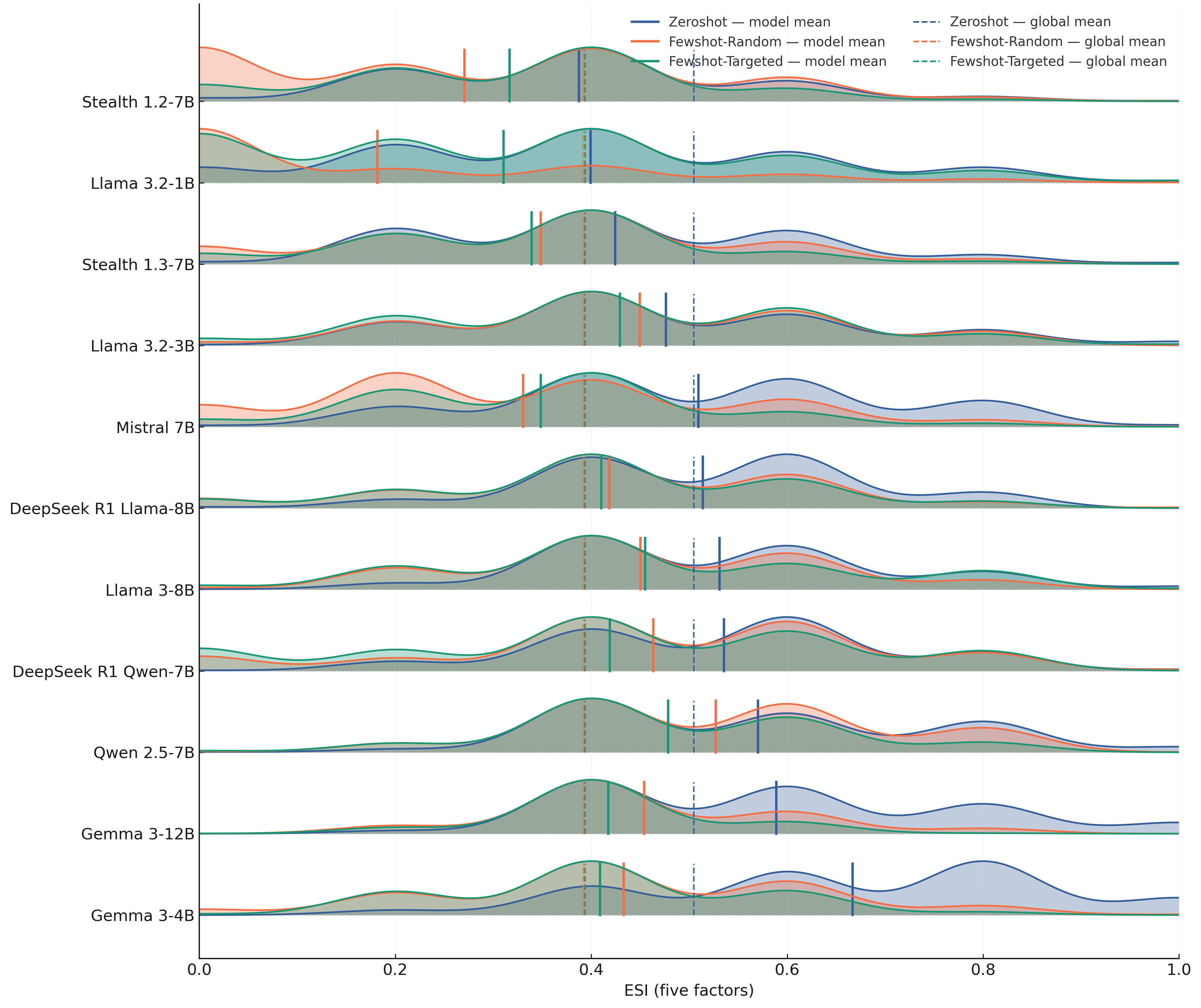}
\caption{Explanation Strength Index Distribution Per Shot Per Model}
\label{fig:sm_esi_dist}
\end{figure}

Factor coverage, measured by the five-factor Explanation Strength Index (ESI) over {time, cost, comfort, convenience, frequency}, shows systematic shifts with shot type (Figure~\ref{fig:sm_esi_dist}). Zeroshot produces the highest means and the widest spreads: global mean ESI lies around $0.55$--$0.60$, consistent with longer, list-like narratives that enumerate several factors. Few-shot prompts reduce ESI to approximately $0.40$--$0.43$ on average and narrow the distributions, which reflects shorter, case-specific rationales once exemplars anchor the relevant trade-offs. The vertical reference lines in Figure~\ref{fig:sm_esi_dist} show that this shift holds across models, although the magnitude varies. Gemma 3 12B, Gemma 3 4B and Qwen 2.5 7B retain relatively high zeroshot ESI and then move to tighter, lower-ESI bands under few-shot while delivering strong accuracy. Stealth 1.2 7B exhibits comparatively low zeroshot ESI and the narrowest targeted curves; the model tends to issue compact, conservative justifications, which matches its calibrated probability profiles. Llama 3.2 1B shows the largest separation between zeroshot and few-shot curves and the lowest targeted mean, consistent with concise but often under-decisive statements.

\begin{figure}[h]
\includegraphics[width=0.9\linewidth]{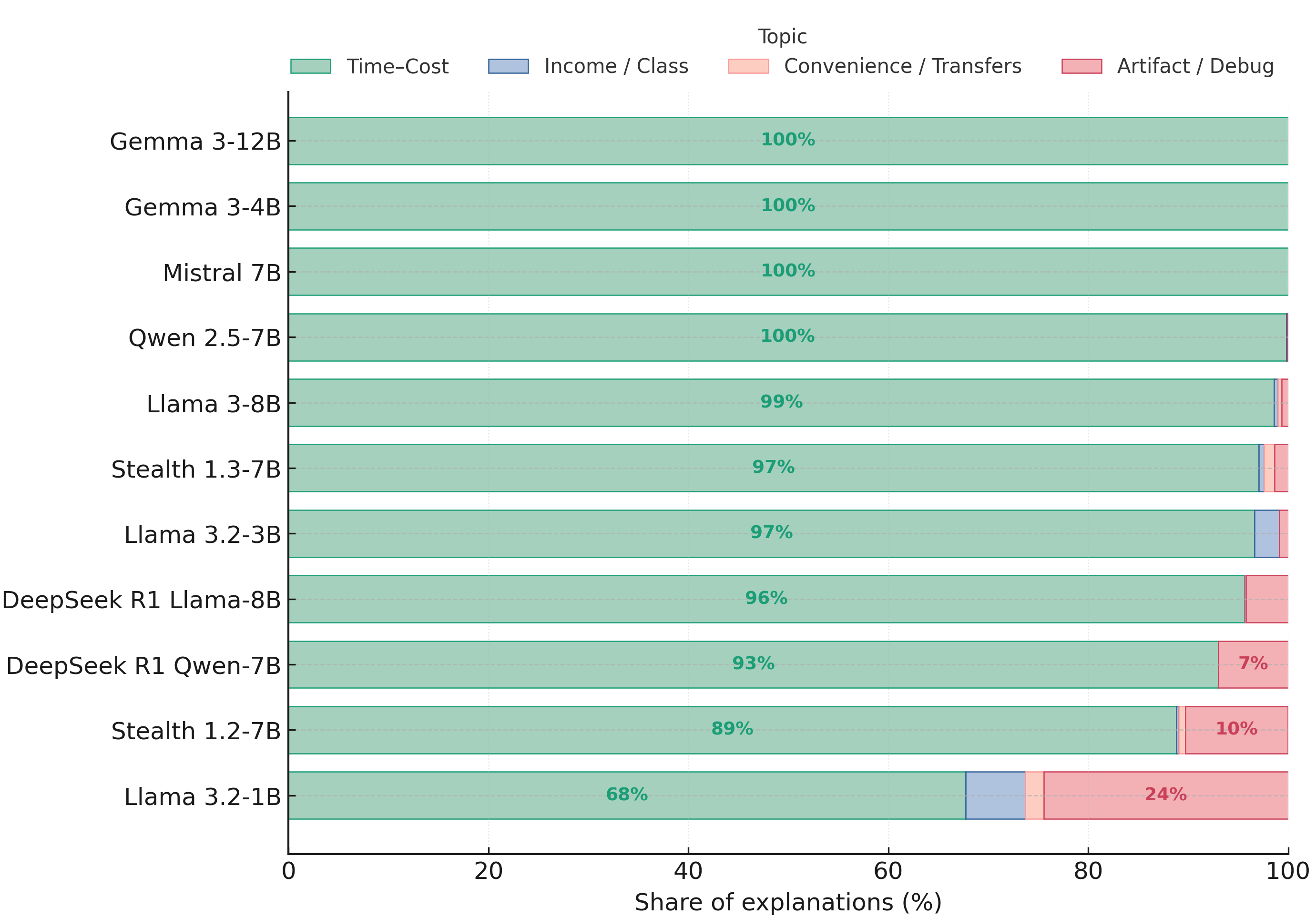}
\caption{Topic Prevalence per Model}
\label{fig:sm_topic_prev}
\end{figure}

Topic prevalence by model, as shown in Figure \ref{fig:sm_topic_prev}, provides a complementary view. Time--Cost dominates across the board but to different degrees. Models that achieve the highest instant-level accuracy devote essentially all text to Time--Cost: Gemma 3 12B sit at $99$--$100\%$. DeepSeek R1 Llama 8B and DeepSeek R1 Qwen 7B remain strongly focused ($\approx 96\%$ and $\approx 93\%$ respectively) but include small Artifact / Debug fractions ($\approx 4\%$ and $\approx 7\%$). Stealth 1.2 7B spreads attention more than most ($\approx 89\%$ Time--Cost, $\approx 10\%$ Artifact / Debug), in line with its well-calibrated yet slightly less decisive predictions. The most diffuse mix appears in Llama 3.2 1B ($\approx 68\%$ Time--Cost, $\approx 6\%$ Income / Class, $\approx 2\%$ Convenience / Transfers, $\approx 24\%$ Artifact / Debug); this profile matches its lower F1 but good distributional fit, since conservative, multi-cue narratives tend to restore probability to minority modes. Aggregated over shots, Income / Class and Convenience / Transfers remain small but interpretable (roughly $1$--$2\%$ and $<1\%$ respectively). 

These textual patterns reconcile the metric trade-offs reported in the pairwise analysis. Statements that focus tightly on Time--Cost align with confident, single-mode selections; weighted F1 rises, but probability mass often shifts away from minority alternatives, which lifts JSD and, on rare omissions, cross-entropy. Narratives that incorporate ticket cues or transfer burden allocate non-negligible probability to secondary modes; JSD and cross-entropy fall, while F1 changes only slightly. Targeted prompting moves occupancy toward the decisive sub-regions inside Time--Cost and toward the two satellites, which explains the simultaneous improvement in calibration and the stability (or modest rise) in instant-level accuracy on Swissmetro.

Finally, the comparison across models suggests a division of strengths. Gemma 3 12B, Gemma 3 4B, Qwen 2.5 7B couple near-pure Time--Cost narratives with high F1 and low variance; these are strong choices for per-trip accuracy. Stealth 1.2 7B and, to a lesser degree, DeepSeek R1 Qwen 7B accept a broader mix of cues and produce the lowest divergence metrics with only modest losses in F1, which suits share-forecasting objectives. Llama 3.2 1B exemplifies the calibration-over-accuracy corner: diffuse narratives and low ESI yield good aggregate shares but reduced top-1 performance. These roles align with the dataset-level results and support mixed strategies: select a high-F1 model for recommendation tasks, or a calibrated model for market-share forecasting, and apply targeted prompting to pull both closer to the efficient frontier.

\subsection{Fine-tuned Model Evaluation and Comparison with Existing Literature and Methods}
\label{fine-tune data}
\begin{figure}[h]
\centering
\includegraphics[width=\linewidth]{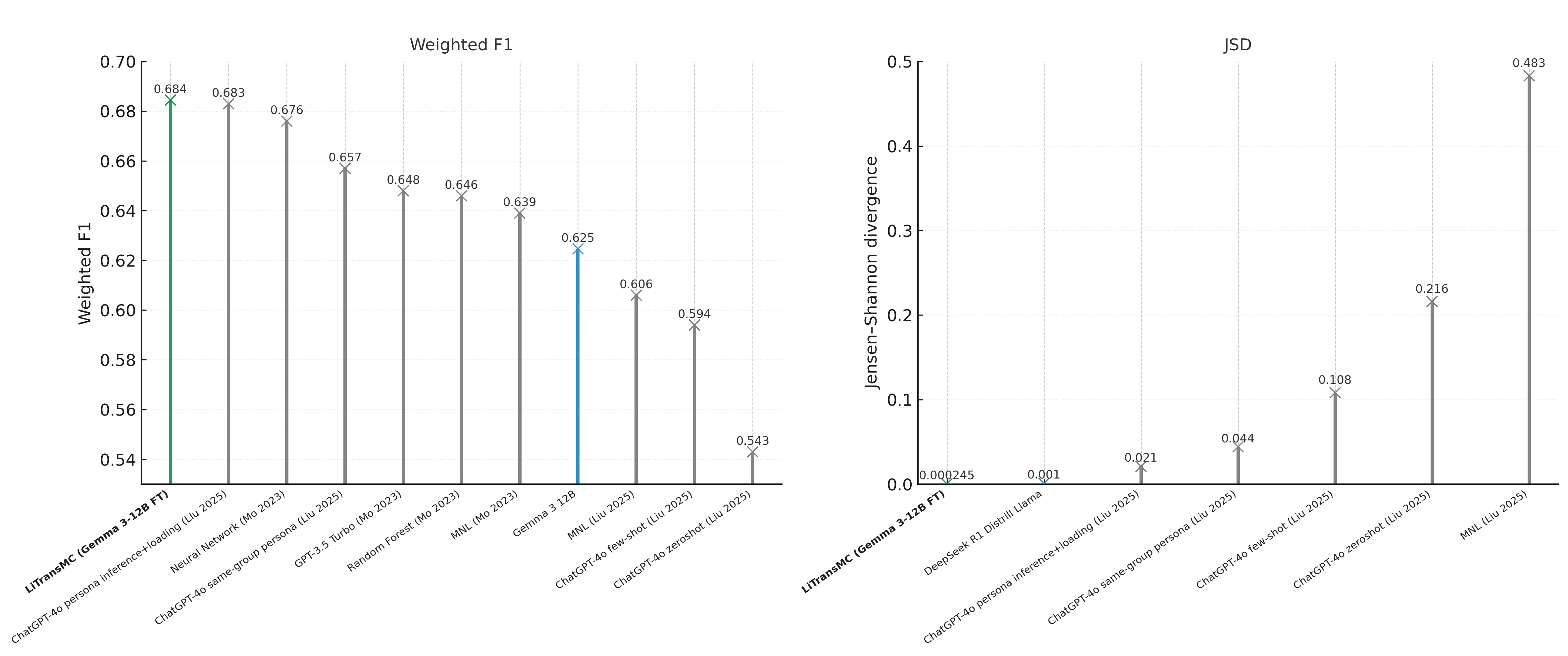}
\caption{LiTransMC (Finetuned model) Performance Evaluation Against Prior Runs and Literature. Left: weighted F1, Right: JSD}
\label{fig:littrans_two_panel}
\end{figure}

We fine-tuned Gemma 3 12B on Swissmetro and refer to the resulting model as ``LiTransMC''. Its performance is summarized in Figure~\ref{fig:littrans_two_panel}, which compares the model against the best untuned local baselines and published results on the same dataset, Swissmetro, from previous studies using proprietary large-scale models (Chat GPT) with different techniques to align LLM behaviour for mode choice experiments as well as discrete choice and machine learning modelling techniques for mode choice modelling on the same dataset using the same data split for training and testing as well as same evaluation metrics.

On individual prediction, LiTransMC under targeted few‑shot attains a weighted F1 of 0.6845. This exceeds the best untuned local run (Gemma 3 12B, targeted few‑shot, 0.6246) and surpasses all GPT baselines from literature on the same dataset: GPT 3.5 Turbo at 0.648 \citep{mo2023large} and the GPT 4o variants in \cite{liu2025aligning}, namely zero‑shot 0.543 and similarity based few‑shot 0.594, as well as the behavioural aligning efforts using targetted few shot with same‑group persona variant at 0.657. It also outperforms classical models reported by \cite{mo2023large} neural network 0.676, random forest 0.646, and MNL 0.639, and the MNL baseline in \cite{liu2025aligning} at 0.606. Relative to the strongest prior large‑model result on this dataset that uses GPT 4o with few shot examples utilizing persona inference and loading (0.683; \citep{liu2025aligning}), LiTransMC is on par and even marginally higher while relying on a compact 12B parameter model and straightforward supervised adaptation rather than a two‑stage persona inference and loading pipeline. Beyond accuracy, locally hosted fine‑tuned models offer practical advantages: they run without external APIs, reduce operating cost, keep data on premises for privacy and safety, and make high‑quality behaviour modelling accessible to teams without access to proprietary frontier models.

On distributional alignment, LiTransMC achieves a JSD of 0.000245, which is lower than every comparator. It improves on the best untuned local JSD (0.000986 from a DeepSeek R1 Distill Llama run) by a factor of about four, and it reduces divergence by roughly eighty‑six times relative to \cite{liu2025aligning} best GPT 4o configuration (0.021). The MNL baseline in same study records a JSD of 0.483, underscoring the magnitude of improvement in reproducing aggregate mode shares. Notably, even without fine‑tuning, the best local baselines already surpass GPT 4o in the zero‑shot and few‑shot settings on weighted F1, indicating that carefully prompted, locally hosted models can be competitive before any supervised adaptation.
These results show that a small, domain‑adapted model can meet or exceed the best reported large‑model accuracy while dramatically improving distributional calibration. In practice, LiTransMC provides three benefits: \emph{privacy preserving local inference, very low and stable divergence that mitigates mode omissions, and materially lower cost of deployment}. While careful prompting already brings untuned models close to the efficient frontier, fine‑tuning shifts the frontier outward by improving both instant‑level accuracy and aggregate alignment in a single deployable model.

\section{Recommendations and Implications}
\label{sec: recomendations}

Our experiments point to two primary factors for reliable mode–choice prediction with causal LLMs: \emph{(i) careful model selection and (ii) the choice of learning style.} Variance decomposition shows that these two factors explain the vast majority of performance variation across datasets, while prompt format and decoding temperature contribute comparatively little. This implies that most of the attainable gains come from picking an appropriate model class and aligning the learning setup to the task, rather than from extensive prompt engineering or sampling stochasticity tuning. 

On the learning style, moving from zero-shot to few-shot consistently improves accuracy, but uncurated examples increase run-to-run variability, which in turn will hinder the models reproducibility and reliability for forecasting tasks. Targeted, similarity-based few-shot learning both raises accuracy and narrows dispersion by retrieving demonstrations that are behaviourally close to the test instance using a multi-component similarity score over socio-demographics, trip attributes, and additional context, rather than relying on cosine similarity alone in matching the most relevant examples for few-shot prompting. In practice, this stabilizes predictions while improving mean performance and should be the default few-shot strategy for travel behaviour modelling. 

Prompt engineering matters, but more as finishing steps than primary drivers. Across models and datasets, direct prompts generally outperform CoT/ReAct once demonstrations are available, and a conservative temperature performs best; raising temperature tends to yield negligible or negative effects. A simple two-step procedure works well in practice: \emph{first, fix a low temperature and choose between direct versus reasoning-style prompts based on validation; second, only then probe minor temperature changes if needed.} 

For the researcher community, these findings suggest the following practical workflow. Start by screening a small set of compact, open models that are known to be strong for this task family, and evaluate them under your target learning style; This step will dominate your returns. Then enable targeted few-shot using structured similarity over the variables that drive behavioural substitution in your context (for example, travel time and cost in the numeric component, exact matching for nominal attributes, and ordinal proximity for ordered factors). This similarity formulation proved more behaviourally aligned than embedding-based cosine approaches on structured survey data. Researchers could also experiment with the different weights attributed to the different components of the similarity score based on their domain knowledge and population heterogeneity. It is also recommended to use direct prompting at low temperature as your initial default, adjusting only at the last stage if you see robust validation gains. 

For the practitioners, two deployment implications stand out. First, revealed-preference settings provide stronger baselines and more stable per-trip predictions than stated-preference settings; when possible, prime the model with RP exemplars and use SP data as an auxiliary source with targeted few-shot to counter hypothetical bias. Second, compact open models that are lightly fine-tuned on domain data can match or exceed proprietary APIs while preserving privacy and controlling cost through local inference, making routine behavioural analytics more accessible. 

When fine-tuning is feasible, a parameter-efficient pass on an appropriate open model shifts the frontier outward: instant-level accuracy improves beyond strong untuned baselines and aggregate alignment tightens substantially, while eliminating dependence on external APIs. This makes a fine-tuned, locally hosted classifier a practical default for agencies and consultancies that must handle sensitive travel data. 

The reasoning analysis adds important operational lessons. Targeted few-shot tightened explanations around the dominant time–cost trade-off while preserving a purposeful role for secondary cues such as transfers or ticketing, which helped protect minority mode probability mass. We therefore recommend reporting a compact reasoning card alongside predictions during development, then constraining explanations in production to a brief statement of the top drivers while archiving full rationale, prompt and exemplars for audit. Future work can formalize simple guardrails on explanation quality by setting thresholds on our reasoning diagnostics, for example, an Explanation Strength Index above 0.60, with runs flagged when this is violated. In parallel, researchers can test sensitivity to exemplar content by adding or removing specific cues such as transfers or ticketing and then measuring the resulting shifts in explanation topics and aggregate mode shares.

More broadly, this work aligns with the evolution towards the broader framework of agentic, human-in-the-loop transportation analytics. A specialized mode-choice LLM can act as a ``choice agent'' within an agentic transportation system: retrieving behaviourally similar exemplars, producing decisions with calibrated probabilities, explaining trade-offs in plain language, and handing off to downstream simulation or optimization tools for verification. The LiTransMC pipeline demonstrates that such specialized, privacy-preserving models are viable today and can serve as foundational building blocks for more interactive, human-centred planning workflows.

\section{Conclusions}
\label{sec:conclusions}

This study systematically evaluated eleven locally deployable, open-access LLMs for transport mode choice prediction and showed that reliable performance depends mainly on two design decisions: the choice of base model and the learning style. Prompt template and decoding temperature offered only marginal, last-mile effects relative to those primary levers. Moving from zero-shot to curated few-shot improved accuracy and stability across datasets, and a similarity-based approach to exemplar selection was consistently beneficial.

A central contribution is LiTransMC, a domain-adapted causal LLM fine-tuned with a parameter-efficient strategy. LiTransMC achieved a weighted F1 of 0.6845 and a Jensen–Shannon divergence of 0.000245, outperforming untuned local baselines and larger proprietary references reported in prior work, while narrowing the gap between instance-level accuracy and aggregate calibration. The results indicate that specialized, locally hosted models can meet or exceed the performance of general-purpose systems while preserving privacy and controlling cost.

Across datasets, a clear pattern emerged: models that are strong in zero-shot provide stable baselines, models that benefit most from high-quality demonstrations reach the highest peaks under curated few-shot, and compact learners improve markedly when examples are well matched. Revealed-preference data produced stronger baselines than stated-preference data, which suggests that empirically grounded decisions provide a cleaner learning signal; this points to opportunities to reduce hypothetical bias in future survey designs.

Collectively, this work provides a proof of concept for transforming general purpose, open-source models into specialized, sovereign AI assets for public agencies. Locally hosted and fine-tuned on confidential survey data, such models offer privacy preserving, cost-effective, and auditable tools capable of supporting nuanced, evidence driven transport planning. By bridging predictive accuracy, distributional fidelity, and interpretability, this study establishes that conversational LLMs are not merely viable alternatives but are poised to become superior instruments for transport behavioural modelling. Future research should focus not on whether these models can be applied, but on how best to operationalize and scale them as bespoke, domain specific AI instruments for strategic policy design and simulation.

\section*{Authorship contribution statement}
The authors confirm their contribution to the paper as follows: \textbf{Tareq Alsaleh}.: Conceptualization, Methodology, Data curation, Investigation, Formal analysis, Software, Visualization, Writing - original draft, and Writing - review \& editing. \textbf{Bilal Farooq}.: Conceptualization, Methodology, Investigation, Funding acquisition, Project administration, Resources, Supervision, and Writing - review \& editing.

\section*{Acknowledgements}
We would like to express our gratitude to Dr. Mohesn Nizami and Dr. Nael Alsaleh, who conducted the original Stated Preference Survey with Brightwater buyers in October 2022 and for making their data available for the validation of the LLMs predictive performance. 
This research was funded by a grant from the Canada Research Chair program in Disruptive Transportation Technologies and Services (CRC-2021-00480) and NSERC Discovery (RGPIN-2020-04492) fund.

\section*{Declaration of generative AI use}
Locally deployed large language models were the subject of investigation in this paper using a methodology, protocol, and analysis entirely developed by the authors. AI use was confined to the research experiment as described in the manuscript. For manuscript preparation, AI tools (e.g., Grammarly, ChatGPT) were used only for grammar and language refinement, with no AI-generated original content. The authors reviewed and edited the content as needed and take full responsibility for the content of the published article.
\newpage
\bibliographystyle{abbrvnat}
\singlespacing
\bibliography{References.bib}
\end{document}